\documentclass[table]{article} 
\usepackage[accepted]{tmlr}

\usepackage{amsmath,amsfonts,bm}

\def\eqref#1{equation~\ref{#1}}

\def\1{\bm{1}}

\DeclareMathAlphabet{\mathsfit}{\encodingdefault}{\sfdefault}{m}{sl}
\SetMathAlphabet{\mathsfit}{bold}{\encodingdefault}{\sfdefault}{bx}{n}

\usepackage[utf8]{inputenc} 
\usepackage[T1]{fontenc}    
\usepackage{microtype}
\usepackage{minted,graphicx,url} 
\usepackage[algo2e,ruled,vlined,linesnumbered]{algorithm2e}
\usepackage{amsmath,amsthm,amssymb}
\usepackage{url}
\usepackage{booktabs}
\usepackage{array}
\usepackage{placeins}
\usepackage{multicol}
\usepackage{tabularx,longtable}
\usepackage{comment}
\usepackage{multirow}
\usepackage{ifthen}
\usepackage{wrapfig}

\usepackage{algorithm}
\usepackage{algorithmic}

\usepackage{newfloat}
\usepackage{listings}
\usepackage{colortbl}

\usepackage{amsmath}
\usepackage{amssymb}
\usepackage{pifont}
\usepackage{xspace}
\usepackage{mathtools}
\usepackage{amsthm}
\usepackage{booktabs}
\usepackage{paralist}
\usepackage{caption}
\usepackage{subcaption}
\usepackage{tabularx}

\usepackage[textsize=tiny]{todonotes}
\usepackage{multirow}
\usepackage{tabularx}

\usepackage[most,skins,theorems]{tcolorbox}

\usepackage{xr}
\usepackage[bookmarksnumbered,debug,backref=page]{hyperref}   
\usepackage[capitalise,nameinlink]{cleveref}

\crefname{table}{Tab.}{Tabs.}
\crefname{section}{Sec.}{Secs.}
\crefname{appendix}{App.}{Apps.}

\graphicspath{%
  {figures/}
}

\makeatletter
\newcommand\myref[1]{
    \@ifundefined{r@#1}{
        \cref*{ext:#1}
    }{
        \cref{#1}
    }
}
\makeatother

\newcolumntype{C}{>{\centering\arraybackslash}X}
\newcolumntype{R}{>{\raggedleft\arraybackslash}X}
\newcolumntype{L}{>{\raggedright\arraybackslash}X}

\newcommand{\cmark}{\ding{51}\xspace}%
\newcommand{\cmarkg}{\textcolor{lightgray}{\ding{51}}\xspace}%
\newcommand{\xmarkg}{\textcolor{lightgray}{\ding{55}}\xspace}%

\newcommand{\moe}{MoE}
\newcommand{\MixtureOfExperts}{Mixture-of-Experts}
\newcommand{\imagenetonek}{ImageNet-1k}
\newcommand{\imagenettwok}{ImageNet-21k}

\newcommand{\alessandro}[1]{\textcolor{black}{#1}}
\newcommand{\olivier}[1]{\textcolor{black}{#1}}

\newcommand{\eg}{e.g.,}

\tcbset{
  aibox/.style={
    width=\linewidth,
    top=8pt,
    bottom=4pt,
    colback=blue!6!white,
    colframe=black,
    colbacktitle=black,
    enhanced,
    center,
    attach boxed title to top left={yshift=-0.1in,xshift=0.15in},
    boxed title style={boxrule=0pt,colframe=white,},
  }
}
\newtcolorbox{AIbox}[2][]{aibox,title=#2,#1}
\definecolor{lightblue}{rgb}{0.22,0.45,0.70}

\definecolor{changecolor}{rgb}{1,0,0} 
\definecolor{normalcolor}{rgb}{0,0,0} 

\newif\ifshowchanges
\showchangesfalse 
\newcommand{\change}[1]{%
  \ifshowchanges
    {\begingroup\color{changecolor}#1\endgroup}%
    \else
    {\begingroup\color{normalcolor}#1\endgroup}%
  \fi
}

\newcommand{\ConvNeXt}{\change{ConvNeXt}}

\definecolor{linkcolor}{RGB}{83,83,182}

\hypersetup
{
  colorlinks=true,
  citecolor=linkcolor,
  linkcolor= linkcolor,
  urlcolor=blue,
  pdfstartview=FitH,
  pdftitle={},
  pdfauthor={},
  pdfcreator={},
  pdfsubject={},
  pdfkeywords={},     
}

\everymath{\displaystyle}

\newcommand*{\hurl}[1]{\href{https://#1}{#1}}

\author{\name Mathurin Videau \email mvideau@meta.com \\
      \addr Meta AI, TAU, INRIA, and LISN (CNRS \& Univ. Paris-Saclay)
      \AND
      \name Alessandro Leite \email aleite@insa-rouen.fr \\
      \addr INSA Rouen Normandy, University of Rouen Normandy, LITIS UR 4108
      \AND
      \name Marc Schoenauer \email marc.schoenauer@inria.fr\\
      \addr TAU, INRIA and LISN (CNRS \& Univ. Paris-Saclay)
      \AND
      \name Olivier Teytaud \\
      \addr
      Thales, CortAIx-Labs}

\def\month{10}  
\def\year{2025} 
\def\openreview{\url{https://openreview.net/forum?id=hKise4AJgp}} 

\begin{document}

\frenchspacing

\title{Mixture of Experts for Image Classification: What's the Sweet Spot?}

\maketitle




\begin{abstract}
\MixtureOfExperts{}~(\moe{}) models have shown promising potential for parameter-efficient scaling across domains. However, their application to image classification remains limited, often requiring billion-scale datasets to be competitive. In this work, we explore the integration of~\moe{} layers into image classification architectures using open datasets. We conduct a systematic analysis across different~\moe{} configurations and model scales. We find that moderate parameter activation per sample provides the best trade-off between performance and efficiency. However, as the number of activated parameters increases, the benefits of~\moe{} diminish. 
\change{Our analysis yields several practical insights for vision \moe{} design. First, \moe{} layers most effectively strengthen tiny and mid-sized models, while gains taper off for large-capacity networks and do not redefine state-of-the-art ImageNet performance. Second, a \textit{Last-2} placement heuristic offers the most robust cross-architecture choice, with \textit{Every-2} slightly better for Vision Transform (ViT), and both remaining effective as data and model scale increase. Third, larger datasets (e.g., ImageNet-21k) allow more experts, up to 16, for \ConvNeXt\ to be utilized effectively without changing placement, as increased data reduces overfitting and promotes broader expert specialization. Finally, a simple linear router performs best, suggesting that additional routing complexity yields no consistent benefit.}
\end{abstract}

\section{Introduction}

Recent advances in machine learning, particularly in the domains of natural language processing~(NLP)\alessandro{\citep{vaswani2017attention, kenton:19bert}} and computer vision~\citep{dosovitskiy:20}, have been primarily driven by scaling up model size, computational budgets, and training data. Although these large-scale models demonstrate impressive performance, they are often expensive to train and consume considerable energy resources~\citep{strubell:19}. As a result, the research community has become increasingly interested in exploring more efficient training and serving paradigms. One such promising solution is the use of sparse expert models, with \MixtureOfExperts{}~(\moe)~\citep{shazeer2017outrageously, lepikhin2020gshard} emerging as a popular variant.

\moe{} models introduce sparsity by partitioning the set of parameters into multiple parallel sub-models, called ``experts.'' During training and inference, the gating part of the models routes input examples to specific expert(s), ensuring that each example only interacts with a subset of the network parameters. As the computational cost is partially correlated with the number of parameters activated for a given sample rather than the total number of parameters, this approach facilitates the scaling-up of the model, while keeping computational costs under control, making it an attractive option for a wide range of applications.

Despite their success in various domains~\citep{hihn2021mixture, costa2022no, zoph2022designing, fedus2022switch}, the application of \moe{} models in image classification and computer vision, in general, remains limited, and it often requires very large datasets~\citep{riquelme2021scaling, mustafa2022multimodal, komatsuzaki2022sparse} to be competitive against state-of-the-art approaches. In this work, we focus on leveraging the potential of~\moe{} models for image classification on~\imagenetonek{} and \imagenettwok~\citep{russakovsky:imagenet:15}. We study the efficiency of integrating \moe{} within two renowned architectures, ConvNext~\citep{liu2022convnet} and Vision Transformer (ViT)~\citep{touvron2022deit}. We conduct a series of experiments considering various architecture configurations. Likewise, we investigate the impact of various components, including the number of experts and their sizes, the gate design, and the layer positions, among others. Our experimental findings indicate that optimal design is contingent on the specific network architecture, and that consistently situating the \moe{} layer within the final two even blocks invariably yields substantial improvements for moderate model size. Nevertheless, when scaling up the approach to large models and datasets close to the state-of-the-art, the benefits of using \MixtureOfExperts{} for image classification gradually vanish.

\paragraph{Contributions.}
In this work, we provide a systematic study of applying \MixtureOfExperts{} (\moe{}) models to image classification tasks, focusing on the ImageNet-1k and \imagenettwok{t} benchmarks. \change{Training follows supervised image classification in two setups, either on ImageNet-1K or ImageNet-22K}. Unlike prior efforts that emphasize large-scale deployments with vast compute resources, we explore \moe{} integration within mid-sized \ConvNeXt{} and Vision Transformer (ViT) architectures, identifying effective design principles for efficient training and inference.
\change{Our analysis yields the following practical guidance for vision \moe{} design:
\begin{itemize}
    \item \textbf{When MoE helps (scale):} \moe{} strengthens tiny and mid-sized models, but gains diminish for large-capacity networks, and do not redefine state-of-the-art ImageNet performance.
    \item \textbf{Placement heuristic:} A \textit{Last-2} configuration is the most robust cross-architecture choice, with \textit{Every-2} slightly better for ViT. These placements remain effective when scaling data or model size.
    \item \textbf{Scaling with data:} With larger datasets (e.g., ImageNet-21k), more experts can be used effectively, up to 16 for ConvNeXt, without changing placement. Increased data reduces overfitting and enables broader expert utilization.
    \item \textbf{Routing choice:} A simple linear router works the best, indicating that added routing complexity brings no consistent benefit.
\end{itemize}
}
%

\section{Related Work}
The~\MixtureOfExperts{}~(\moe{}) model, introduced by~\citep{jacobs1991adaptive}, partitions complex tasks into hopefully simpler sub-tasks handled by expert models, whose predictions are combined to produce the final output. This framework has been successfully integrated into various neural network architectures, particularly transformers for NLP tasks~\citep{shazeer2017outrageously}. Given this success in NLP, interest has increased in applying \moe{} to the diverse and complex computer vision tasks.
 
The transformative potential of \moe{} architectures is underscored by their successful integration across various domains, with the Vision Transformer (ViT) being a prime example. \citet{riquelme2021scaling} introduced V-MoE, an \moe{}-augmented ViT model, for image classification tasks on a massive dataset containing hundreds of millions of examples. They showed that V-MoE not only matches the performance of prior state-of-the-art architectures but also requires half the computational resources during inference. \Citet{lou2021cross} presented a sparse \moe{} MLP model based on the MLP-Mixer architecture~\citep{tolstikhin2021mlp}. While the MLP-Mixer architecture does not display performance accuracy, the \moe{}-enhanced version surpassed its dense counterpart in experiments on ImageNet and CIFAR. \citet{hwang2022tutel} demonstrated the effectiveness of SwinV2-MoE, a MoE-based model built upon Swin Transformer V2 architecture. They reported superior accuracy in downstream computer vision tasks. Similarly, \citet{puigcerver2023sparse} introduced a soft~\moe{} mechanism, demonstrating improved performance and training speed compared to classical~\moe{} on billion-scale datasets.
\change{More recently, \citet{han2024vimoe} proposed ViMoE, which investigates expert placement strategies by fine-tuning DINOv2 models. In contrast, our work focuses on training \moe{} architectures from scratch, enabling a systematic study of their behavior across both ConvNeXt and ViT backbones. This distinction allows us to provide complementary insights into how \moe{} mechanisms contribute to compute efficiency and performance in vision models.}

\section{Sparsely Activated \moe{}}\label{sec:moe}
Based on conditional computations, \moe{} aims at activating specific parts of the model depending on the input. The core idea is to assign experts to different regions of the input space, thereby increasing the model capacity by augmenting the number of parameters without incurring significant computational overhead~\citep{jacobs1991adaptive}.

According to~\citet{shazeer2017outrageously}, an \moe{} includes a router $G$ and a set of experts $E$. The router learns a sparse assignment between the input and the experts, while the experts process the inputs like standard neural network modules do. Let $x$ be the input, and let  $(E_i)_{i \in [1, N]}$ be the   $N$ experts of the \moe{} layer. Following~\citep{shazeer2017outrageously}, the output of this \moe{} layer is given by:
%
\begin{equation*}
    \sum_{i\in \text{Top}_k(x)} G(x)_i \cdot E_i(x)
\end{equation*}
where, $G$ is a conv1x1 gate employing a softmax function, and $G(x)_i$ is its $i^{th}$ output. To promote sparsity in the~\moe{} layer, the number of participating experts is restricted to $k<N$ for each input: $\text{Top}_k(x)$ contains the indices of the $k$ highest $G(x)_i$.
Note that involving $k > 1$ experts increases the computational cost compared to the sparse ($k=1$) counterpart. We tested several gate designs, but none of them clearly outperformed the simple conv1x1 gate~(\Cref{sec:routing:architecture}). 

However, enforcing this \moe{} design is insufficient to yield an effective \moe{}. To exploit compute capacity optimally, we encourage uniform usage of each expert by using a load balancing auxiliary loss proposed in~\citep{shazeer2017outrageously}, which promotes balanced expert assignments and equal importance of experts~(for more details, see Section 4 of~\citet{shazeer2017outrageously}). We also use Batch Prioritized Routing (BPR) introduced by~\citep{riquelme2021scaling}.


\subsection{Vision Transformer}\label{sec:vit}


\begin{figure*}
\footnotesize{%
    \begin{subfigure}[c]{.47\textwidth}
      \includegraphics[width=\columnwidth]{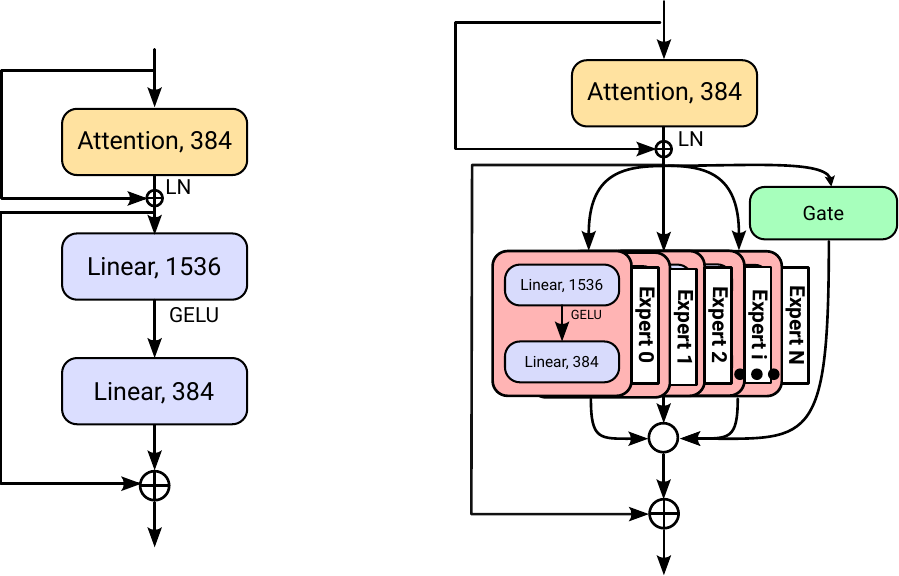}
       \caption{\alessandro{\textbf{Left}}: ViT base block. \textbf{Right}: Our modified \olivier{\moe{}} ViT block. LN: Layer normalization. GELU: Gaussian Error Linear Units.
      }\label{fig:moe_vit_block}
    \end{subfigure}
    \hspace*{1em}
    \begin{subfigure}[c]{.47\textwidth}
        \includegraphics[width=\columnwidth]{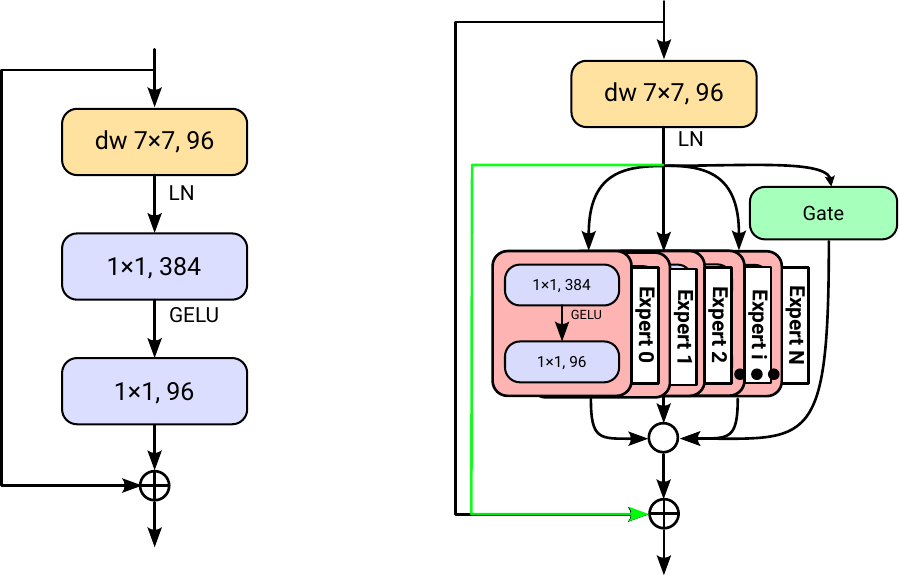}
        \caption{\textbf{Left}: \ConvNeXt\ original base block. \textbf{Right}: The modified \moe{} \ConvNeXt\ block. The green link represents the added skip connection. DW: Depth-wise. LN: Layer normalization. 
    }
    \label{fig:block}
    \end{subfigure}%
    \caption{Vision Transformer and ConvNext architectures.}
}
\end{figure*}

The Vision Transformer (ViT)\alessandro{~\citep{dosovitskiy:20}} represents a paradigm shift in computer vision. It moves away from traditional convolution-based architectures in favor of the Transformer's attention mechanism, initially pioneered for natural language processing tasks~\citep{vaswani2017attention}. In the ViT model, an image is segmented into fixed-size and non-overlapping patches. These patches are linearly embedded into vectors and then passed through a series of Transformer blocks. Each Transformer block is predominantly made up of two main components: a multi-head self-attention mechanism and a multi-layer perceptron (MLP). Within the context of ViT, the MLP ratio (denoted mlp\_ratio in the following) refers to the ratio of the hidden dimensions of the MLP to the embedding dimensions. This ratio is pivotal, as it determines the capacity and computational cost of the MLP component. We replace some predefined MLP block by a~\moe, as depicted in~\cref{fig:moe_vit_block}. Each \moe{} operates at the patch level: the experts are spatially distributed. \change{We consider \begin{inparaenum}[(a)] \item a standard ViT architecture~(ViT-B (Base)) and \item a variant with 8 experts (ViT-B-8).\end{inparaenum}}

\subsection{ConvNeXt}\label{sec:convnext}
%
ConvNeXt~\citep{liu2022convnet} has modernized the classical convolutional neural networks~(CNN) design by incorporating insights from Vision Transformers (ViT). The central component of the ConvNeXt architecture is a block containing a $7 \times 7$ depth-wise convolution followed by a $1\times 1$ convolutional layer (\cref{fig:block}-left). This block is replicated across four stages, each with decreasing resolution. 
In this work, we replace some predefined ConvNeXt block, namely the multilayer perceptron, with a \moe{} block, and introduce a skip connection, as illustrated in~\cref{fig:block}-right. Consequently, each expert becomes a fully convolutional block operating at the feature map level, resulting in spatially distributed experts. Hence, this design not only increases the network capacity but also allows each expert to specialize in specific spatial locations and engages multiple experts per image. \change{We examine three variants: \begin{inparaenum}[(a)] \item ConvNeXt-T (Tiny), ConvNeXt-S (Small), and ConvNeXt-B (Base). These architectures primarily differ in depth and width; and \item ConvNeXt-S-4 and ConvNeXt-B-4, which are variants with 4 experts.\end{inparaenum}} 

\section{Experiments on ImageNet}\label{ImageNetExperiments}
%
This section presents the results obtained by the architectures described above (that incorporate~\moe{} into \ConvNeXt\ and ViT architectures),  
trained on~\imagenetonek{} or pre-trained on \imagenettwok{} datasets. It includes the experimental setup, the results on the \imagenetonek{} validation set, and the impact of various \moe{} configurations.
\subsection{Experimental Setup}\label{sec:xpsetup1}
All the results presented in this section rely on \moe{} models trained on the ImageNet datasets. Although the context is different due to the changes in network structures presented in~\cref{sec:vit,sec:convnext}, we set the training hyperparameters similar to~\citep{touvron2022deit} for ViT and~\citep{liu2022convnet} for \ConvNeXt\ (a grid search looking for better settings did not bring significant improvement).

Furthermore, when working with the \imagenetonek{} dataset, we use a strong data-augmentation pipeline, including Mixup~\citep{zhang2017mixup}, Cutmix~\citep{yun:19:ccv}, RandAugment~\citep{cubuk:20randaugment}, and Random Erasing~\citep{zhong2020random}, over 300 epochs. Likewise, we utilize drop path, weight decay, and expert-specific weight decay as regularization strategies. 
However, for the pretraining phase on \imagenettwok{}, we exclude Mixup and Random Erasing from the data augmentation pipeline because they did not improve the results, as also reported in \citep{touvron2022deit, tu2022maxvit}. We pre-train the model for 90 epochs, adhering to the original \ConvNeXt\ approach~\citep{liu2022convnet} and, consistently with \citep{liu2022convnet} and \citep{touvron2022deit}, the final results of \imagenetonek{} are obtained through fine-tuning for 30 epochs for \ConvNeXt\ and 50 for ViT. Comprehensive details of all the hyperparameters are provided in~\cref{tab:hparams} in~\cref{sec:detailedhypers}.

\subsection{Base results on ImageNet}\label{sec:Im:res}

\begin{table*}[t]
    \centering    
    \scriptsize{
    \caption{Accuracy for different ImageNet-1K trained models and the \moe{} strategies ``Every 2'' and ``Last 2'' (see text). 
    ``Top $k$'' corresponds to the number of experts involved. 
    Throughput is measured on V100 GPUs, following \citep{touvron2021training}. For~\imagenetonek{}, non-isotropic \ConvNeXt\ models feature an mlp\_ratio of 2, which contributes to their improved flops-per-sample efficiency compared to their dense counterparts. 
    }\label{tab:1k:reduced}
    \begin{tabularx}{\linewidth}{lCCCCC}
        \toprule
        \multicolumn{1}{c}{\multirow{2}{*}{Architecture}} & \verb|#|Params & {\small Per sample} & \multicolumn{1}{c}{\multirow{2}{*}{FLOPs}} & Throughput & IN-1K \\
         & ($\times10^6$) & \verb|#|Params$_{act}$  &  & (im/s) & Accuracy\\ 
        \hline
        \rowcolor{gray!15} ConvNeXt-T \citep{liu2022convnet} & 28.6 & 28.6 & 4.5G & 814 & 82.1\\
        \rowcolor{gray!15} ConvNeXt-T-4 Last 2 Top 1 & 34.5 & 25.6 & 4.2G & 768 & 82.1\\
        \cmidrule{1-6}
        ConvNeXt-S \citep{liu2022convnet} & 50 & 50 & 8.7G & 466 & 83.1 \\
         ConvNeXt-S-4 Last 2 Top 1 & 56.1 & 47.3 & 8.5G & 442 & 83.1\\
        \cmidrule{1-6}
        \rowcolor{gray!15} ConvNeXt-B \citep{liu2022convnet}  & 88.6 & 88.6 & 15.4G & 299 & \textbf{83.8}\\
        \rowcolor{gray!15} ConvNeXt-B-4 Last 2 Top 1& 99.1 & 83.4 & 15.0G & 289 & 83.5\\
        \cmidrule{1-6}
        \rowcolor{gray!15} ConvNeXt-S \textit{(iso.)} & 22.3 & 22.3 & 4.3G & 1100 & 79.7\\
        \rowcolor{gray!15} ConvNeXt-S-8 \textit{(iso.)}  Last 2 Top 1 & 38.9 & 22.3 & 4.3G & 1031 & \textbf{80.3}\\
        \cmidrule{1-6}
        ConvNeXt-B \textit{(iso.)} & 82.4 & 82.4 & 16.9G & 336 & \textbf{82.0}\\
        ConvNeXt-B-8 \textit{(iso.)} Last 2 Top 1 & 115.4 & 82.4 & 16.9G & 303 & 81.6\\
        \cmidrule{1-6}
        \rowcolor{gray!15} ViT-S & 22.0 & 22.0 & 4.6G & 1083 & 79.8\\
        \rowcolor{gray!15} ViT-S-8 Last 2 Top 2 & 38.6 & 25.0 & 5.3G & 892 & 80.5\\
        \rowcolor{gray!15} ViT-S-8 Every 2 Top 2 & 71.7 & 33.1 & 6.9G & 724 & \textbf{80.7}\\
        \cmidrule{1-6}
        ViT-B & 86.6 & 86.6 & 17.5G & 329 & \textbf{82.8}\\
         ViT-B-8 Every 2 Top 2 & 284.9 & 129.9 & 26.3G & 227 & 82.5\\
        \bottomrule
    \end{tabularx}    
}
\end{table*}

\begin{table*}[t]
    \centering    
    \scriptsize{%
    \caption{Accuracy for ImageNet-21K pre-trained models and different adaptations with ImageNet-1K: {\em IN-1K} is standard fine-tuning, {\em Linear prob} is the training of a linear layer for the output probability layer (everything else being frozen), and {\em 0-shot} is the direct application of the pre-trained model to ImageNet-1K. ``Every 2'' and ``Last 2'' are the corresponding \moe{} strategies (see text). 
    ``Top $k$'' corresponds to the number of experts involved. 
    }\label{tab:21k:reduced}
    \begin{tabularx}{\linewidth}{lCCCCCC}
        \toprule
        \multicolumn{1}{c}{\multirow{2}{*}{Architecture}} & \verb|#|Params & {\small Per sample} & \multicolumn{1}{c}{\multirow{2}{*}{FLOPs}} & IN-1K & Linear prob & 0-shot\\
         & ($\times10^6$) & \verb|#|Params$_ {act}$ &  & Accuracy & Accuracy & Accuracy\\ 
         \hline
       \rowcolor{gray!15} ConvNeXt-T \citep{liu2022convnet} & 28.6 & 28.6 & 4.5G & 82.9 & 81.8 & 44.3 \\
        \rowcolor{gray!15} ConvNeXt-T-8 Last 2 Top 1 & 70.0 & 28.7 & 4.5G &\textbf{83.5} & \textbf{82.3} & \textbf{44.6} \\
        \hline
        ConvNeXt-S \citep{liu2022convnet} & 50.3 & 50.3 & 8.7G & 84.6 & 83.2 & 45.2 \\
        ConvNeXt-S-8 Last 2 Top 1 & 91.6 & 50.3 & 8.7G & \textbf{84.9} & \textbf{83.6} & \textbf{45.3}\\
        \hline
        \rowcolor{gray!15} ConvNeXt-B \citep{liu2022convnet} & 88.6 & 88.6 & 15.4G & \textbf{85.8} & \textbf{84.9} & 45.8 \\
        \rowcolor{gray!15} ConvNeXt-B-8 Last 2 Top 1 & 162.0 & 88.6 & 15.4G & 85.7 & 84.8 & \textbf{45.9} \\
        \hline
        ViT-S &  22.0 & 22.0 & 4.6G & 82.6 & 81.7 & 44.2 \\
         ViT-S-8 Every 2 Top 2 &  71.7 & 33.1 & 6.9G & \textbf{83.0} & \textbf{81.9} & \textbf{44.7}\\
        \hline
        \rowcolor{gray!15} ViT-B &  86.6 & 86.6 & 17.5G & 85.2 & 84.0 & \textbf{45.7} \\
        \rowcolor{gray!15} ViT-B-8 Every 2 Top 2 & 284.9 & 129.9 & 26.3G & 85.2 & 84.0 & 45.6\\
        %
        \bottomrule
    \end{tabularx}
    }    
\end{table*}

\Cref{tab:1k:reduced} presents the results obtained on \imagenetonek{} validation set by a model that has been entirely trained on \imagenetonek, for isotropic architecture~(\eg{} ViT, \ConvNeXt\ \textit{iso.}) and a hierarchical architecture, namely \ConvNeXt{}. We see some significant improvements in accuracy compared to the ``no-\moe{}'' results for small model sizes, especially for anisotropic models. 

\begin{figure}[!htb]
    \centering
    \begin{subfigure}[c]{.45\textwidth}\centering
        \includegraphics[width=\textwidth]{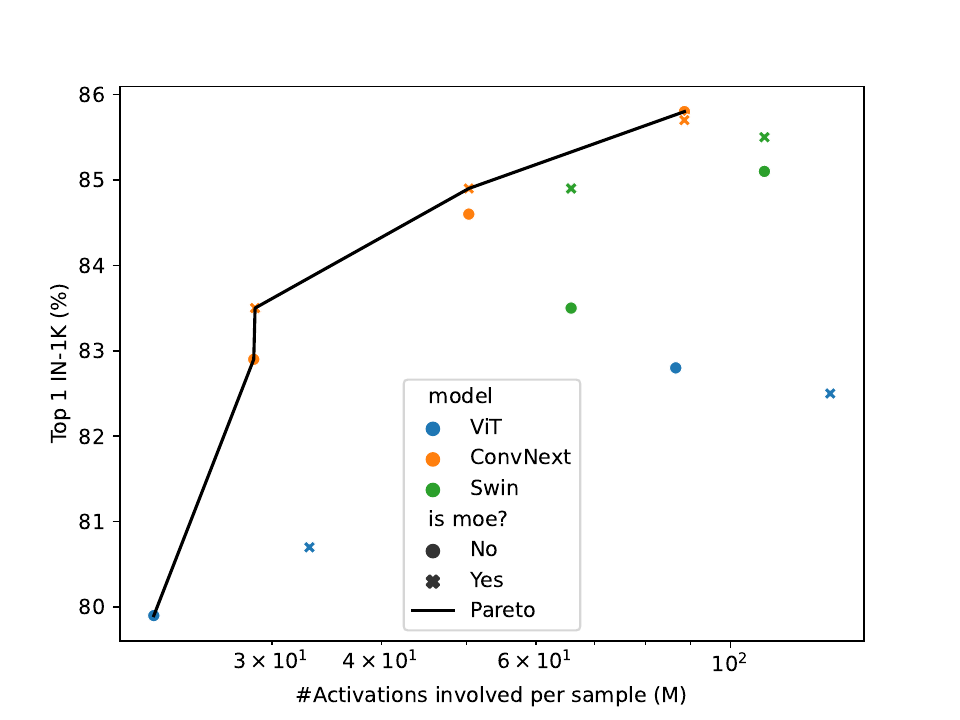}
    \end{subfigure}
    \begin{subfigure}[c]{.45\textwidth}
        \centering
        \includegraphics[width=\textwidth]{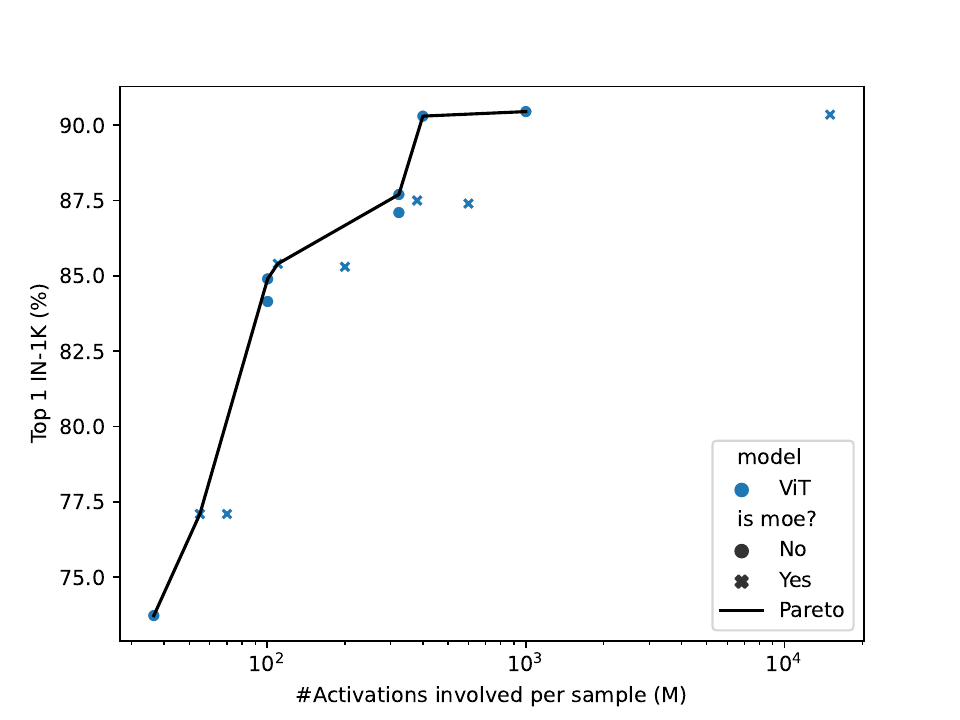}
    \end{subfigure}
    \caption{\olivier{\textbf{Left}: Pareto-front for ImageNet21k, x-axis = number of activations per sample. ViT models have been presented in \moe{} versions only after additional pretraining, and are therefore not presented. \moe{} seems to be Pareto optimal for a number of activations per sample below 90M. \textbf{Right}: Pareto-front for ViT models on JFT-300M. Overall, \moe{} is never validated for a number of activations per sample above $\approx$ 100M.} }
    \label{fig:pfs}
\end{figure}

\Cref{tab:21k:reduced} presents the results of models that are pre-trained on \imagenettwok{}, and tested on the same \imagenetonek{} validation set than above. Here, \moe{} does bring some improvement for moderate numbers of activations per sample. However, this improvement decreases for large numbers of activations per sample, as depicted in~\cref{fig:pfs}.

\begin{AIbox}{\moe{} Strengthens Moderate Model Size; Not the Frontier}
    \moe{} integration provides clear gains at a moderate scale, with tiny and small models benefiting most. As data scale grows, these relative improvements strengthen. Yet as model capacity rises, these benefits diminish, showing that \moe{} enhances mid-sized models but does not redefine SoTA performance.
\end{AIbox}

\subsection{Sensitivity with Respect to Design Choices}\label{sec:abla}
%
This section presents different sensitivity studies that investigate the specifics of the design choices of the~\moe{} layers. We explore the impact of the position of these layers inside the architecture~(\cref{sec:abla:placement}), the effect of varying the number of experts~(\cref{sec:abla:nb_exp}),
and finally, we assess the necessary design changes when transitioning from the~\imagenetonek{} to \imagenettwok{} dataset~(\cref{sec:abla:data}) and the impact of the routing architecture choices~(\cref{sec:routing:architecture}).\change{Ideally, sensitivity analyses would explore the joint effects of multiple factors (such as the number, placement, and size of experts), but exhaustive multi-axis exploration quickly becomes computationally intractable. Nevertheless, we partially address such interactions by analyzing the combined impact of expert size and expert count in~\cref{sec:abla:size_number}, highlighting how these factors jointly influence efficiency and accuracy.}
\subsubsection{Impact of the Position of \moe{} Layers}\label{sec:abla:placement}

%

\begin{table}[!h]
    \centering 
{
\caption{\footnotesize{Comparative results for different positions of \moe{} layers: ImageNet-1k training on \ConvNeXt-T and ViT-S, all employing 8 experts.}}\label{tab:1kplacement}
\scriptsize
\begin{tabularx}{\linewidth}{
 r
 >{\hsize=.30\hsize}C
 >{\hsize=.14\hsize}C
 >{\hsize=.14\hsize}C
 >{\hsize=.14\hsize}C
 >{\hsize=.25\hsize}C
}
  \toprule
& & \multicolumn{2}{c}{\# Params} & \\
\cmidrule{3-4}
\multicolumn{1}{c}{Architecture} & \multicolumn{1}{c}{MoE} & $\times10^6$ & Sample & \multirow{1}{*}{FLOPs} & IN-1K \\
\midrule
\rowcolor{gray!15} \shortstack{ConvNext-T} & {\bf{no \moe}} & 28.6 & 28.6 & 4.5G & {\bf{82.1}}\\
\cmidrule{1-6}
\rowcolor{gray!15}  & every 2 & 54.3 & 17.0 & 3.8G & 81.8\\
\rowcolor{gray!15}  & {\bf{stage}} & 47.4 & 25.5 & 4.0G & {\bf{82.1}}\\
\rowcolor{gray!15}  & {\bf{last 3}} & 49.9 & 25.0 & 4.1G & {\bf{82.1}}\\
\rowcolor{gray!15} \multirow{-4}{*}{ConvNext-T-8} & {\bf{last 2}} & 46.3 & 25.6 & 4.2G & {\bf{82.1}}\\
 \cmidrule{1-6}
 \shortstack{ConvNeXt-S \\\textit{(iso.)}} & \multirow{-2}{*}{no moe} & \multirow{-2}{*}{22.3} & \multirow{-2}{*}{22.3} & \multirow{-2}{*}{8.7G} & \multirow{-2}{*}{79.7}\\
 & every 2 & 96.7 & 22.3 & 8.7G & 79.6\\
 \multirow{-2}{*}{\shortstack{ConvNeXt-S-8 \\ \textit{(iso.)}}} & {\bf{ last 2}} & 38.9 & 22.3 & 8.7G & \textbf{80.3}\\
 \cmidrule{1-6}
 \rowcolor{gray!15} ViT-S & no \moe & 22.0 & 22.0 & 4.2G & 79.9\\
 \rowcolor{gray!15} & {\bf{every 2}} & 71.7 & 33.1 & 5.3G & \textbf{80.7}\\
 \rowcolor{gray!15} \multirow{-2}{*}{ViT-S-8} & last 2 & 38.6 & 25.0 & 6.9G & 80.5\\
\bottomrule
ı\end{tabularx}
}
\end{table}
%
%

\cref{tab:1kplacement} displays the effects of various placements of \moe{} layers, comparing the following configurations:
\begin{inparaenum}[(a)]
  \item \textbf{Every 2}: a \moe{} layer replaces every second layer; 
  \item \textbf{Stage}: a \moe{} layer replaces the final layer of each stage, resulting in four \moe{} layers throughout the network for \ConvNeXt{}; %
  \item \textbf{Last 2}: a \moe{} layer replaces the final layer of each of the last two stages; %
  \item \textbf{Last 3}: Same as ``Last 2'', but with an additional \moe{} layer inserted in the middle of stage 3.
\end{inparaenum}
%
%
%
As shown in~\cref{tab:1kplacement}, ``Last 2'' strategy is the most robust choice: it performs well across all architectures. In contrast, ``Every 2'' performs worst for \ConvNeXt\ architecture, and the best for ViT. Consequently, in the following, all experiments use ``Last 2'' for \ConvNeXt\ and ``Every 2'' for ViT.  

\begin{AIbox}{\moe{} placement matters}
    While ViT benefits most from inserting \moe{} layers at every second block and \ConvNeXt\ from placing them in the last two stages, the Last 2 strategy offers a simple, broadly effective rule of thumb that consistently delivers strong performance across architectures.
\end{AIbox}

\subsubsection{Influence of the Number of Experts}\label{sec:abla:nb_exp}
%
%
%

\begin{table}
   \begin{minipage}[t]{0.49\linewidth}
      \centering      
         \centering 
\scriptsize
{%
\caption{
Comparative results for different numbers of experts. ImageNet-1k training using the ``Last 2'' strategy. 
}\label{tab:1knumexp}
\scriptsize
\begin{tabularx}{\linewidth}{rcCCCCC}
  \toprule
  \multicolumn{1}{c}{\multirow{4}{*}{Architecture}} & \multicolumn{1}{c}{\multirow{4}{*}{\# experts}} & {\multirow{4}{*}{Top-k}} &\verb|#|Params & Per samples & IN-1K \\
  & & & ($\times10^6$) & \verb|#|Params$_{act}$   & Top 1 acc.\\ 
\midrule
\cmidrule{1-5}
  \rowcolor{gray!15}  \ConvNeXt-T & \textbf{no \moe{}} & - & 28.6 & 28.6 & \textbf{82.1}\\
\cmidrule{1-6}
  \rowcolor{gray!15}   & {\bf{4}} & 1 & 54.3 & 25.5  & {\bf{82.1}}\\
  \rowcolor{gray!15}   & {\bf{8}} & {\bf{1}} & 47.4 & 25.5  & {\bf{82.1}}\\
  \rowcolor{gray!15}   & 8 & 2 & 47.4 & 25.5  & 82.0\\
  \rowcolor{gray!15}  \multirow{-4}{*}{\ConvNeXt-T} & 16 & 1 & 46.3 & 25.5 & 81.7\\
 \cmidrule{1-6}
 \cmidrule{1-6}
 ViT-S & no \moe & - & 22.0 & 22.0 & 79.9\\
 \cmidrule{1-6}
  & 4 & 2 & 29.1 & 25.0 & 79.8\\
  & 8 & 1 & 38.6 & 22.0 & 80.2\\
  & {\bf{8}} & {\bf{2}} & 38.6 & 25.0 & {\bf{80.5}}\\
 \multirow{-4}{*}{ViT-S} & 16 & 2 & 57.5 & 25.0 & 80.2\\
\bottomrule
\end{tabularx}
}
   \end{minipage}
   \hspace{0.001\linewidth}
   \begin{minipage}[t]{0.49\linewidth}
     \centering      
     \centering
\scriptsize
{%
\caption{ImageNet-21k training on \ConvNeXt-T employing 8 experts and an MLP ratio of 4 unless specified explicitly.
}\label{tab:21k:placement}
\begin{tabularx}{1\linewidth}{
   >{\hsize=.21\hsize}C
   >{\hsize=.17\hsize}C
   >{\hsize=.22\hsize}C
   >{\hsize=.11\hsize}C
   >{\hsize=.27\hsize}C
   }
  \toprule
  \multirow{2}{*}{MoE} & \verb|#|Params & Per samples & \multirow{2}{*}{FLOPs} & IN-1K \\
     strategy & ($\times10^6$) & \verb|#|Params$_{act}$  &  & Top 1 acc.\\ 
  \midrule
  no \moe & 28.6 & 28.6 & 4.5G & 82.9\\
\rowcolor{gray!15}
every 2 & 97.5 & 28.6 & 4.5G & 82.9\\
{\bf{stage}} & 78.2 & 28.6 & 4.5G & {\bf{83.5}}\\
\rowcolor{gray!15}
{\bf{last 3}} & 70.0 & 28.6 & 4.5G & {\bf{83.5}}\\
{\bf{last 2}} & 70.0 & 28.6 & 4.5G & {\bf{83.5}}\\
\rowcolor{gray!15}
last 2 mlp\_ratio 2 & \multirow{3}{*}{46.3} & \multirow{3}{*}{25.6} & \multirow{3}{*}{4.2G} & \multirow{3}{*}{83.4}\\
\bottomrule
\end{tabularx}
}
   \end{minipage}
\end{table}

\Cref{tab:1knumexp} presents the performance of ConvNext and ViT architectures on ImageNet-1K training with the ``Last 2'' strategy \moe{} layers. It provides a comprehensive view of how the number of experts influences the size (parameter count) of the models and their Top-1 accuracy on this dataset.

%
The experimental results suggest that four experts yield the best performance for \ConvNeXt{}, while eight experts are optimal for ViT. This is showcased by the Top-1 accuracy rates, which are consistent at 82.1\% for \ConvNeXt\ with four experts and reach a peak of 80.5\% for ViT with eight experts. However, we note that increasing the number of experts to 16 has a detrimental effect on both architectures. Specifically, for \ConvNeXt{}, the top-1 accuracy slightly drops to 81.7\%, and for ViT, the performance plateaus at 80.2\%.


These findings highlight that while \moe{} layers can enhance model performance, there is a delicate trade-off to be struck in terms of the number of experts. Exceeding the optimal number can lead to suboptimal results, negating the potential benefits of the \moe{} integration.  Furthermore, in our detailed analysis, described in~\cref{sec:abla:sizevsnum},
we explore the interplay between the number and the size of each expert. Our investigation revealed that, for isotropic networks such as ViT and \ConvNeXt\ \textit{iso.}, reducing the size of experts adversely affects performance, while for \ConvNeXt{}, it has no significant impact. Moreover, during the course of our experiments, we observed that while the Top-1 configuration was superior for \ConvNeXt{}, the Top-2 configuration for ViT was comparable or even better than the \ConvNeXt\ Top-1 configuration, leading us to employ Top-1 for \ConvNeXt\ and Top-2 for ViT \change{(cf. \Cref{tab:1knumexp})}.
\begin{AIbox}{Expert Count under Moderate Data Scale}
    On ImageNet-1k, ConvNeXt models reach peak performance with 4 experts, while ViT benefits from up to 8. This contrast stems from architectural width: ConvNeXt's broad final stages saturate quickly, limiting the value of additional experts, whereas ViT's narrower last stage tolerates more experts.
\end{AIbox}
\subsubsection{Impact of Expert Size and Number}\label{sec:abla:sizevsnum}
\begin{figure*}[!ht]
    \scriptsize
    \begin{subfigure}[c]{.44\textwidth}
       \includegraphics[width=\textwidth]{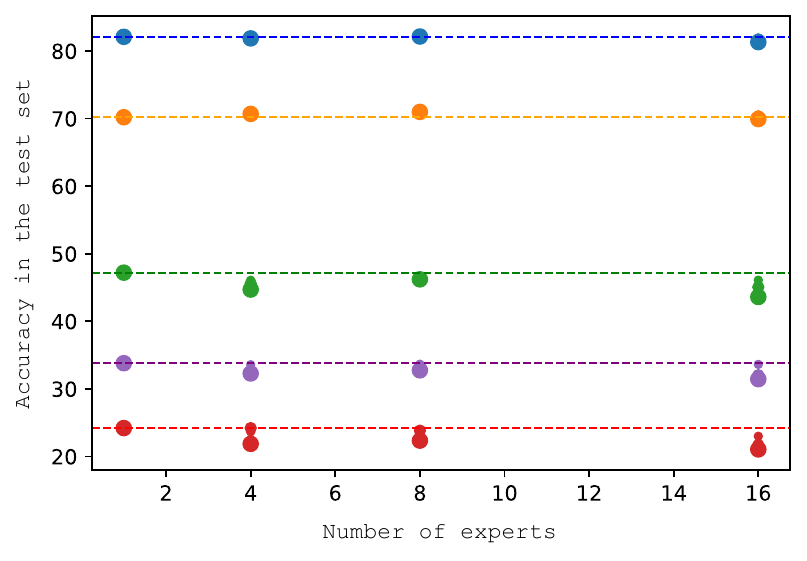}
       \caption{\# experts ($\in\{1,4,8,16\}$) vs. performance. \textbf{Points' size}: MLP-ratio ($\in\{1,2,4\}$).}
    \end{subfigure}
    \hspace{5pt}
    \begin{subfigure}[c]{.54\textwidth}
        \includegraphics[width=\textwidth]{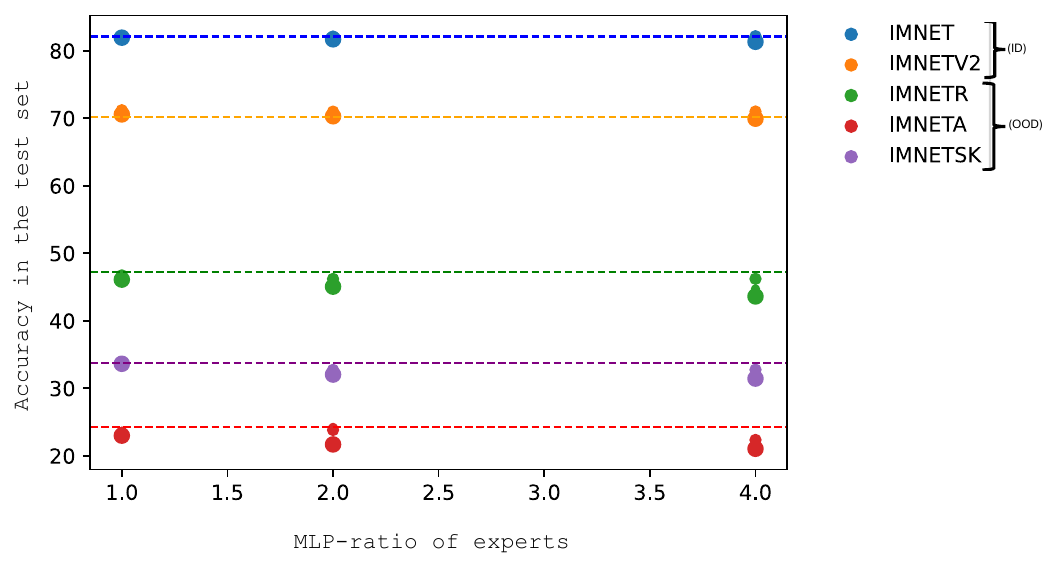}
        \caption{Experts' MLP-ratio ($\in \{2,4\}$) vs. performance. \textbf{Points' size}: \# experts ($\in \{1,4,8,16\}$).}
    \end{subfigure}
      \caption{Exploring the interplay between the size and count of experts in a \olivier{\moe{}} layer for \ConvNeXt-T on ImageNet-1K. Baseline results (without \moe{}) are denoted by dotted lines. For this small dataset, \moe{} does not bring much improvement (see~\cref{fig:pfs} for bigger datasets).}
    \label{fig:abla:sizevsnum}
\end{figure*}

Two critical tunable parameters in the \moe{} layers are the number of experts per layer and the size of each expert. The latter is defined by the mlp\_ratio, which scales the hidden dimension of the feed-forward network~(FFN)~(\cref{sec:vit}).
\cref{fig:abla:sizevsnum} illustrates the sensitivity of ConvNeXt's performance to variations in these parameters, with the number of experts ranging from 4 to 16 and the mlp\_ratio from 1 to 4. By default (original network), this ratio is set to 4, meaning that values below 4 correspond to smaller per-sample FFN capacities than in the original network.

To assess generalization, we report top-1 accuracy across three dataset categories: (i) in-distribution (ID) datasets—ImageNet and ImageNetV2~\citep{recht2019ImageNet}; (ii) out-of-distribution (OOD) datasets—ImageNet-R~\citep{hendrycks2021many} and ImageNet-Sketch~\citep{wang2019learning}; and (iii) adversarial datasets—ImageNet-A~\citep{hendrycks2021nae}. The objective is to determine whether performance benefits more from using many smaller experts or fewer larger ones.

Our experiments on ConvNeXt reveal that while \moe{} layers can slightly improve or preserve performance on ID datasets, they tend to degrade performance on OOD datasets. In particular, increasing the number of experts to 16 consistently reduces accuracy across all dataset types. Regarding expert size, smaller experts appear marginally more robust to increases in expert count. Overall, the best tradeoff for ID is achieved with 4 to 8 experts and an mlp\_ratio of 2 or 4. Based on these findings, we adopt a configuration of 4 experts with an mlp\_ratio of 2 for experiments on ImageNet-1K. For ConvNeXt isotropic and ViT architectures, however, reducing the mlp\_ratio consistently harms performance, and in the case of ViT, it also introduces training instabilities. Also, OOD performance is typically degraded in~\cref{fig:abla:sizevsnum}.

\begin{AIbox}{Balancing Expert Number and Width on ImageNet-1k}
    On ImageNet-1k, the best trade-off is achieved with 4–8 experts and mlp\_ratio 2–4; increasing to 16 experts consistently reduces accuracy and OOD robustness. Increasing number of experts while reducing their size helps reducing overfitting.
\end{AIbox}
\subsubsection{Results on \imagenettwok{}}\label{sec:abla:data}
%

\cref{tab:21k:placement,tab:abla:21kexp} present the results of models trained on \imagenettwok{}, reporting \imagenetonek{} accuracy after fine-tuning. 
We analyse the effect of \moe{} layer positioning, expert size (\cref{tab:21k:placement}), and expert count~(\cref{tab:abla:21kexp}). The results demonstrate that, when large volumes of data are accessible, a greater number of experts can be effectively deployed. Notably, the use of sixteen experts, which was sub-optimal for ImageNet-1k, does not negatively affect performance on \imagenettwok{}. This suggests that increasing the number of experts, together with the volume of data, could lead to valuable enhancements.

In terms of \moe{} layer configuration, our data showed no performance improvements with the addition of more layers, and the ``every 2'' strategy again yielded the poorest results for \ConvNeXt-based architecture. 

\begin{AIbox}{Scaling Expert Count with Larger Pretraining}
    On ImageNet-21k, ConvNeXt scales up to 16 expert,s and ViT also benefits from more. Increased data reduces overfitting, allowing more experts. This highlights the link between dataset scale and MoE capacity. The optimal layer positioning remains unchanged.
\end{AIbox}
\begin{table}
  \vspace{-\baselineskip}
  \centering
\scriptsize
{%
\caption{ImageNet-21k training on \ConvNeXt-T: comparison between various numbers of experts, with an MLP ratio of 4.}\label{tab:abla:21kexp}
\begin{tabularx}{\linewidth}{
  >{\hsize=.22\hsize}C
  >{\hsize=.22\hsize}C
  >{\hsize=.27\hsize}C
  >{\hsize=.13\hsize}C
  >{\hsize=.22\hsize}C
}
 \toprule
\multirow{2}{*}{\# experts} & \verb|#|Params & {Per samples} & \multirow{2}{*}{FLOPs} & IN-1K \\
 & ($\times10^6$) & \verb|#|Params$_{act}$  &  & Top 1 acc.\\ 
\midrule
  \rowcolor{gray!15}  no \moe{}   & 28.6 & 28.6 & 4.5G & 82.9\\
\hline
8 & 54.3 & 28.6 & 4.5G & 83.5\\
  \rowcolor{gray!15}  {\bf{16}} & 117.1 & 28.6 & 4.5G & {\bf{83.6}}\\
32 & 211.6 & 28.6 & 4.5G & 83.4\\
\bottomrule
\end{tabularx}
}
\end{table}
\subsubsection{Impact of the Routing Architectures}\label{sec:routing:architecture}
\begin{wraptable}{r}{.4\linewidth}
\vspace{-\baselineskip}
    \centering     
   \scriptsize{
    \caption{Performance of MoE-Integrated Models vs Non-MoE Baselines: ablation study on the gating network architecture. MoE layers are placed on the last two even layers.}\label{tab:abla:gates}
    \centering
    \textbf{\ConvNeXt-S}\\
    \begin{tabularx}{\linewidth}{ccCCCC}
    \toprule
   \rowcolor{gray!15}  \multicolumn{1}{c}{Gate} & IN-1K & IMV2 & IMA & IMR & IMSK \\
    \hline
    no \moe & \textbf{82.1} & 70.8 & {24.2} & \textbf{47.2} & \textbf{33.8}\\
    \cmidrule{1-6}
     \rowcolor{gray!15}  Conv & \textbf{82.1} & \textbf{71.0} & 23.8 & 46.2 & 32.7\\
    Cos	& 81.9 & 70.6 & 22.3 & 45.6 & 33.0\\
    \rowcolor{gray!15}   L2	& 82.0 & 71.6 & \textbf{25.0} & 45.9 & 32.8\\
    \cmidrule{1-6}
    \end{tabularx}\\
    \centering
    \textbf{\ConvNeXt-S \textit{(iso.)}}\\
     \begin{tabularx}{\linewidth}{ccCCCC}
   \rowcolor{gray!15}    \multicolumn{1}{c}{Gate} & IN-1K & IMV2 & IMA & IMR & IMSK \\
    \cmidrule{1-6}
    no \moe & 79.7 & 68.6 & 13.0 & 46.4 & 34.0\\
    \cmidrule{1-6}
  \rowcolor{gray!15}     Conv & \textbf{80.3} & \textbf{68.8} & \textbf{13.5} & \textbf{46.6} & \textbf{34.4}\\
    Cos	& 79.7 & 68.4 & 12.2 & 45.3 & 33.0\\
   \rowcolor{gray!15}    L2	& 80.1 & 68.6 & 13.3 & 46.1 & 34.3\\
    \cmidrule{1-6}
    \end{tabularx} \\ 
    \centering
    \textbf{ViT-S}\\
     \begin{tabularx}{\linewidth}{ccCCCC}
   \rowcolor{gray!15}    \multicolumn{1}{c}{Gate} & IN-1K & IMV2 & IMA & IMR & IMSK \\
    \cmidrule{1-6}
    no \moe & 79.8 & \textbf{69.1} & 19.8 & \textbf{43.4} & 29.7\\
    \cmidrule{1-6}
  \rowcolor{gray!15}     Conv & \textbf{80.5} & \textbf{69.1} & \textbf{20.3} & 43.2 & \textbf{29.9}\\
    Cos	& 79.7 & 67.8 & 15.9 & 41.6 & 29.3\\
   \rowcolor{gray!15}    L2	& 80.2 & 68.8 & 19.0 & 41.9 & 29.2\\
    \bottomrule
 \end{tabularx}
}
\vspace{-60pt}
\end{wraptable}

We investigated three distinct routing architectures for our study:
\begin{itemize}
    \item ``Conv'': A straightforward $1 \times 1$ convolution, similar to a linear gate.
    \item ``Cos'': A gate that utilizes cosine similarity for routing, as detailed in~\citep{chi2022representation}. It is a two-layer architecture, with one linear layer projecting the features to a lower-dimensional space and another performing cosine similarity against some learned latent codes.
    \item ``L2'': This is identical to the ``Cos'' gate, except that it employs the L2-distance for similarity instead of cosine.
\end{itemize}

We conducted training on the ImageNet-1K dataset using various small networks, each employing different routing mechanisms as displayed earlier. The results, as shown in~\cref{tab:abla:gates}, indicate that the simple convolutional (conv) configuration generally yields slightly better performance in most cases. However, when evaluating these models on out-of-distribution (OOD) datasets, the results are more varied. The inclusion of OOD evaluation is crucial, as an effective routing mechanism should ideally facilitate an expert decomposition that generalizes well. Notably, the \moe{} models that showed superior performance on ImageNet-1K also displayed enhanced robustness in the OOD evaluations.

\begin{AIbox}{Simplicity Wins in Routing Architectures}
    Across both ConvNeXt and ViT, a straightforward linear gate achieves the best overall results. More complex routing adds no clear benefit and often underperforms, even on OOD data.
\end{AIbox}

\section{Discussion}\label{sec:discussion}

\subsection{Hierarchical vs Isotropic Models}

\begin{table}
  \centering
\vspace{-\baselineskip}
    \centering
    \scriptsize%
    {%
    \caption{Robustness of \moe{} w.r.t. to domain change, for two different pretrainings, both fine-tuned on ImageNet-1k: Top-1 accuracies on various datasets.}\label{tab:rob}\footnotesize
    \begin{tabularx}{\linewidth}{rCCCCC}
    \toprule
    \multicolumn{1}{c}{Model} & IN-1K & IMV2 & IMA & IMR & IMSK \\
    \midrule
    \multicolumn{6}{c}{\textbf{Imagenet1k trained}}\\
    \cmidrule{1-6}
    \ConvNeXt-T & \textbf{82.1} & 70.8 & \textbf{24.2} & \textbf{47.2} & \textbf{33.8}\\
    \ConvNeXt-T-4 & \textbf{82.1} & \textbf{71.0} & 23.8 & 46.2 & 32.7\\
    \hline
    \rowcolor{gray!15}   \ConvNeXt-S & \textbf{83.1} & \textbf{72.5} & \textbf{31.3} & \textbf{49.6} & \textbf{37.0} \\
   \rowcolor{gray!15}    \ConvNeXt-S-4 & \textbf{83.1} & 72.2 & 30.2 & 49.0 & 37.3\\
    \hline
    \ConvNeXt-B & \textbf{83.8} & \textbf{73.4} & \textbf{36.7} & \textbf{51.3} & \textbf{38.2}\\
    \ConvNeXt-B-4 & 83.5 & 72.8 & 33.9 & 48.6 & 36.6 \\
    \hline
   \rowcolor{gray!15}    ViT-S & 79.9 & 68.8 & 19.8 & 43.4 & 29.7\\
    \rowcolor{gray!15}   ViT-S-8 & \textbf{80.7} & \textbf{70.1} & \textbf{21.1} & \textbf{43.9} & \textbf{30.9}\\
    \hline
    ViT-B & \textbf{82.8} & \textbf{72.1} & \textbf{32.4} & \textbf{51.2} & \textbf{36.9}\\
    ViT-B-8 & 82.5 & 71.5 & 32.0 & 46.6 & 35.2\\
    \hline
   \rowcolor{gray!15}    \ConvNeXt-S \textit{(iso.)} & 79.7 & 68.6 & 13.0 & 46.4 & 34.0\\
   \rowcolor{gray!15}    \ConvNeXt-S-8 \textit{(iso.)} & \textbf{80.3} & \textbf{68.8} & \textbf{13.5} & \textbf{46.6} & \textbf{34.4}\\
    \hline
    \ConvNeXt-B \textit{(iso.)} & \textbf{82.0} & \textbf{71.1} & \textbf{21.2} & \textbf{50.0} & \textbf{38.1}\\
    \ConvNeXt-B-8 \textit{(iso.)} & 81.6 & 70.6 & 19.1 & 48.5 & 35.7\\
    \cmidrule{1-6}
    \multicolumn{6}{c}{\textbf{Imagenet21k pretrained}}\\
    \cmidrule{1-6}
  \rowcolor{gray!15}     \ConvNeXt-T & 82.9 & 72.4 & \textbf{36.2} & 51.1 & 38.5\\
   \rowcolor{gray!15}    \ConvNeXt-T-8 & \textbf{83.5} & \textbf{72.8} & 32.6 & \textbf{51.3} & \textbf{40.8}\\
    \hline
    \ConvNeXt-S & 84.6 & 74.7 & \textbf{44.8} & 57.5 & 43.6\\
    \ConvNeXt-S-8 & \textbf{84.9} & \textbf{75.5} & 44.4 & \textbf{55.7} & \textbf{45.5}\\
    \hline
   \rowcolor{gray!15}    \ConvNeXt-B & \textbf{85.8} & 76.0 & \textbf{54.6} & \textbf{62.0} & \textbf{48.8}\\
   \rowcolor{gray!15}    \ConvNeXt-B-8 & 85.7 & \textbf{76.3} & 51.8 & 59.6 & 48.4\\
    \hline
    ViT-S & 82.6 & 72.6 & \textbf{38.9} & 50.8 & 39.0\\
    ViT-S-8 & \textbf{83.0} & \textbf{72.7} & 35.3 & \textbf{51.0} & \textbf{39.4}\\
    \hline
   \rowcolor{gray!15}    ViT-B & 85.2 & \textbf{76.1} & \textbf{56.0} & \textbf{61.5} & \textbf{46.9}\\
    \rowcolor{gray!15}   ViT-B-8 & 85.2 & 75.4 & 48.1 & 59.3 & 45.9\\
    \bottomrule
    \end{tabularx}
 }
\end{table}

First, one can note that the best \moe{} designs typically apply experts only to the last layers.
ViT operates at the same resolution at each layer after the input images have been cut into patches. This isotropic design implies that \moe{} is applied in layers that are not particularly large. On the other hand, with hierarchical models such as \ConvNeXt{}, applying \moe{} to the last layers considerably increases model size. Indeed, as presented in~\cref{tab:abla:21kexp}, employing eight experts doubles the size of the network despite using only two layers of~\moe{}.

\subsection{Positions and Numbers of \moe{} Layers}
There are two prevalent strategies in the literature concerning the position of \moe{} layer: ``Every 2'' and ``Last 2''. In the context of hierarchical architectures such as \ConvNeXt{}, the ``Every 2'' strategy dramatically increases the number of weights while simultaneously yielding inferior results. However, for ViT ``Last 2'' and ``Every 2'' are two viable strategies. This is in line with the results of V-MoE~\citep{riquelme2021scaling} on large-scale datasets. Overall, our results confirm that ``Last 2'' is a solid starting point, as it consistently yields positive outcomes across all tested architectures. However, it may not always be the optimal choice for every architecture. For instance, in the case of ViT, ``Every 2'' demonstrates better results. 
Regarding the number of experts, we noted that model performance quickly saturates with the number of experts, as reproduced in~\cref{tab:abla:21kexp}, and detailed in~\cref{fig:abla:sizevsnum}.
\subsection{Robustness Evaluation}
%
We evaluated \moe{} models, trained or fine-tuned on \imagenetonek{}, against a range of robustness benchmark datasets, such as ImageNet-A (IMA)~\citep{hendrycks2021nae}, ImageNet-R~(IMR)~\citep{hendrycks2021many}, ImageNet-Sketch~(IMSK)~\citep{wang2019learning}, and ImageNet~V2~(IMV2)~\citep{recht2019ImageNet}, the latter being used as a measure of overfitting.

\Cref{tab:rob} reports Top-1 accuracy for all datasets. %
%
%
It shows that \moe{} models tend to underperform their dense counterparts in \imagenetonek{}. However, we note that \moe{} models are better for the robustness metric; they were actually already better on \imagenetonek{}. So, these results actually emphasize the success of \moe{} when the model is sufficiently small to be improved by a \moe{}, as discussed in the previous section: the domain generalization holds only in the regime in which \moe{} improves the base results.

\subsection{On Data-Augmentation}

As shown in the previous section,  \moe{} is sensitive to the amount of data at hand. This sensitivity stems from the inherent structure of \moe{}, where increasing the number of experts results in each expert operating on a smaller data fraction. Consequently, each expert tends to learn specific data clusters, as discussed by~\citep{chen2022towards}.

\begin{figure*}
  \centering
    \begin{minipage}[t]{0.49\linewidth}
       \centering
       \vspace{0pt}  
       \scriptsize{
        \captionof{table}{ViT-B trained on \imagenetonek{} with different hyperparameters. In particular, for the top row (gray) we follow the training recipe from Deit \citep{touvron2021training} and for the bottom one we follow Deit III~\citep{touvron2022deit}.}\label{tab:data_aug}
         \begin{tabularx}{\linewidth}
            {ccCC}
            \toprule
             \multicolumn{1}{c}{\multirow{2}{*}{Architecture}} & \# Params & {\small Per sample} & IN-1K \\
             & ($\times10^6$) & \#Params$_{act}$   & Acc.\\ 
            \midrule   
           \rowcolor{gray!15} ViT-B-8 & & & \\
           \rowcolor{gray!15} \citep{touvron2021training}  & \multirow{-2}{*}{86.6} & \multirow{-2}{*}{86.6} & \multirow{-2}{*}{81.8}\\
           \rowcolor{gray!15} ViT-B-8 & & & \\
           \rowcolor{gray!15} Every 2 Top 2 & \multirow{-2}{*}{284.9} & \multirow{-2}{*}{129.9} &  \multirow{-2}{*}{\textbf{82.2}}\\
            \cmidrule{1-4}
            ViT-B & & & \\
            \citep{touvron2022deit} & \multirow{-2}{*}{86.6} & \multirow{-2}{*}{86.6} & \multirow{-2}{*}{\textbf{82.8}}\\
            ViT-B-8 & & &\\
            Every 2 Top 2 & \multirow{-2}{*}{284.9} & \multirow{-2}{*}{129.9} & \multirow{-2}{*}{82.5}\\
            \bottomrule
         \end{tabularx}
      }      
   \end{minipage}
   \hfill
    \begin{minipage}[t]{0.49\linewidth}%
      \vspace{0pt}  
       \centering
        \scriptsize{%
         \includegraphics[width=\linewidth]{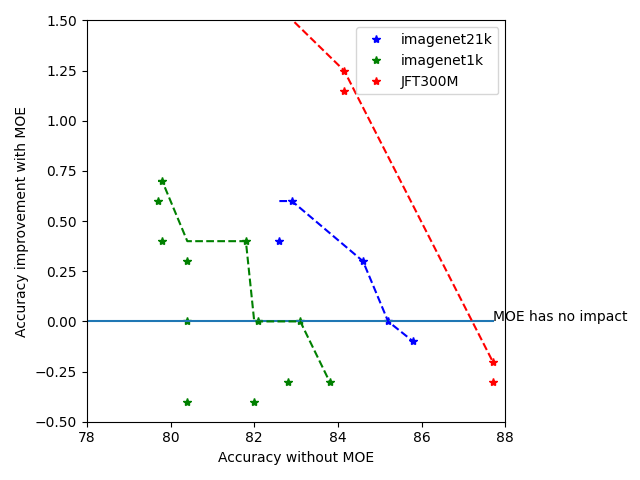}
         \caption{Improvement from~\moe{} (y-axis) vs. baseline accuracy without~\moe{} (x-axis) for image classification, across pretraining sizes. Larger pretraining yields greater gains; higher baseline accuracy reduces impact. Dashed lines show maximal improvement per accuracy level. Results are drawn from~\cref{tab:21k:reduced,tab:1k:reduced,tab:21k:moe}~(in~\cref{app:moedense}).}\label{fig:max_ipvt}
   }%
   \end{minipage}
   \vspace{-0.5em}
\end{figure*}

Modern training methodologies often employ robust data augmentation techniques to enhance model generalization. However, the interplay between these augmentations and the data clusters learned by the \moe{} model remains relatively uncharted. For instance, techniques like Mixup~\citep{zhang2017mixup}, which blend images, could blur the distinctions between these clusters. Similarly, the Random Resize method, with its aggressive magnification capability, may distort the original data clusters. Understanding the precise influence of these augmentations on the \moe{} model's learning process is essential, as it holds implications for the model's ultimate performance. Some evidence of this can be seen in \cref{tab:data_aug}. While one training recipe might show \moe{} outperforming its dense counterparts (as indicated in the top rows of the table), a different, more effective recipe could lead to the opposite outcome (Bottom rows of the table).
%

%

\subsection{Alternate Views}

With the ImageNet benchmark increasingly reaching saturation, the focus in contemporary computer vision models has shifted towards computational efficiency and scaling prowess. An essential question arises: \emph{Does the \moe{} model enhance this aspect in image classification?}
%
Our experiments with ImageNet reveal an intriguing insight: integrating \moe{} into models like \ConvNeXt\ and ViT enhances performance, but this improvement tends to plateau in models with over 100M parameters. At first glance, this might seem to contradict findings from some papers \citep{riquelme2021scaling,mustafa2022multimodal}. However, with additional context, this observation aligns with the existing literature, as evidenced by results in~\cref{fig:max_ipvt,fig:pfs}, and \cref{tab:21k:moe}~(\Cref{app:moedense}). 
Specifically, as the base dense network grows larger, the performance gap between \moe{} and its dense counterpart narrows.

Exploring this further in large-scale datasets, such as JFT 3 billion~\citep{zhai2022scaling}, reveals similar findings. For instance, the vision transformer model SoViT, with 400M parameters~\citep{alabdulmohsin2023getting}, exhibits comparable performance to the 15B V-\moe{}~\citep{riquelme2021scaling} when both use ViT-based architectures. Similarly, when considering the JFT-3B dataset~\citep{zhai2022scaling}, the 5.6B LiMoE-H model~\citep{mustafa2022multimodal} and Coca-Large~\citep{yu2022coca} lead to analogous results when evaluated on a per-sample parameter basis.  Interestingly, in the vision-language modeling domain, \citet{lin2024moma} observe that increasing the number of experts for the image modality does not lead to diminishing returns, aligning with our findings regarding the efficacy of experts in scaling performance.

To summarize, while the \moe{} model does not appear to redefine the state-of-the-art when matched against dense models on both ImageNet and billion-scale datasets, its real strength emerges in enhancing smaller models. By integrating \moe{}, these compact models can achieve better performance, pushing their capabilities closer to the forefront of current benchmarks. Yet, it is essential to acknowledge that, in the broader context, \moe{} might not drive significant advancements in the overall state-of-the-art. \change{Moreover, shape optimization as demonstrated in SoViT, could further improve performance and could even compound with the benefits of \moe{}, enabling smaller and more capable models.} This nuanced perspective is further enforced in the appendix, with the data showcased in~\cref{tab:21k:moe} in~\cref{app:moedense}.

\subsection{Model Inspection}\label{sec:inspect}

\begin{wrapfigure}{r}{0.3\textwidth}
    \vspace{-55pt}
    \centering
    \includegraphics[width=\linewidth]{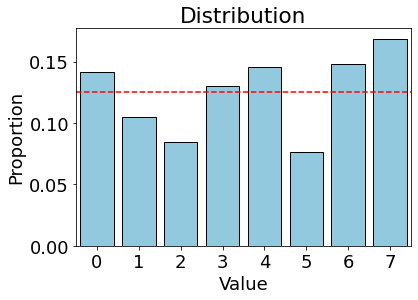}
    \caption{Distribution of experts}\label{fig:dist}
    \vspace{-25pt}
\end{wrapfigure}

The objective of this section is to examine the routing component of the model. We will present the top expert for each routing layer. Note that when considering the top-2 configuration, we only report, for this visualization, the top-1 of these two experts.

\begin{figure}
\scriptsize{%
    \centering
    \vspace{-0.1cm}
    \begin{subfigure}[c]{.195\textwidth}
        \includegraphics[width=\textwidth]{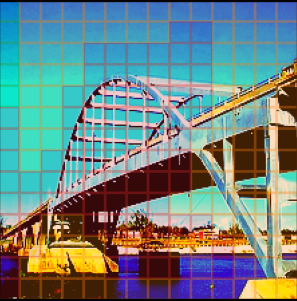}
    \end{subfigure}
    \begin{subfigure}[c]{.195\textwidth}
        \includegraphics[width=\textwidth]{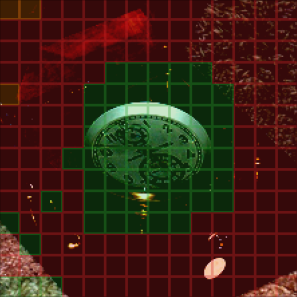}
    \end{subfigure}
    \begin{subfigure}[c]{.195\textwidth}
        \includegraphics[width=\textwidth]{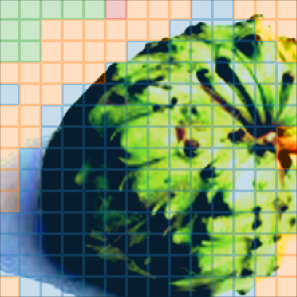}
    \end{subfigure}
    \begin{subfigure}[c]{.195\textwidth}
        \includegraphics[width=\textwidth]{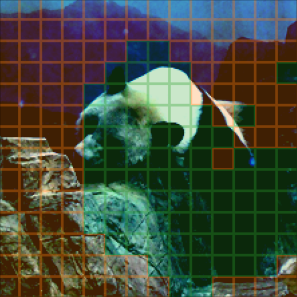}
    \end{subfigure}
    \begin{subfigure}[c]{.195\textwidth}
        \includegraphics[width=\textwidth]{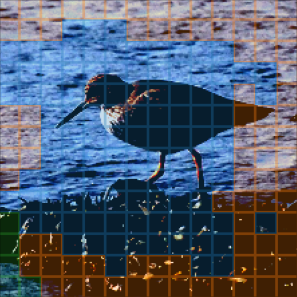}
    \end{subfigure}
    \begin{subfigure}[c]{.195\textwidth}
        \includegraphics[width=\textwidth]{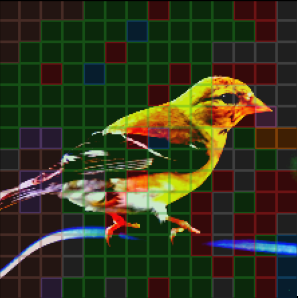}
    \end{subfigure}
    \begin{subfigure}[c]{.195\textwidth}
        \includegraphics[width=\textwidth]{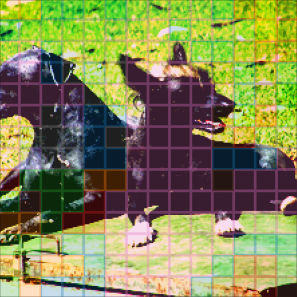}
    \end{subfigure}
    \begin{subfigure}[c]{.195\textwidth}
        \includegraphics[width=\textwidth]{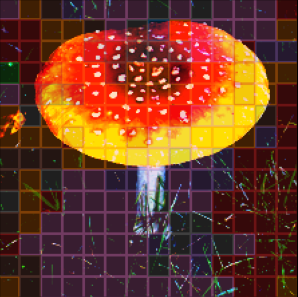}
    \end{subfigure}
    \begin{subfigure}[c]{.195\textwidth}
        \includegraphics[width=\textwidth]{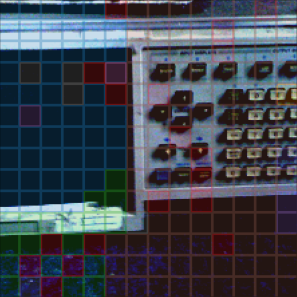}
    \end{subfigure}
    \begin{subfigure}[c]{.195\textwidth}
        \includegraphics[width=\textwidth]{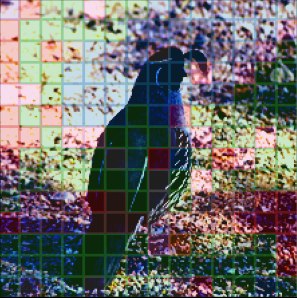}
    \end{subfigure}
    \caption{The top row features the ConvNext-T-4 model, while the bottom row showcases the ViT-S-8 model, both of which have been trained on the ImageNet1k dataset. The images displayed have been generated by upsampling the gating output, retaining only the top expert for clarity. Each distinct color represents a unique expert. These specific images are samples from the ImageNet validation set, extracted from the last third stage of the \ConvNeXt-T-4 and the final layer of the ViT-S-8 models, respectively.}\label{fig:moe_vizu}
}
\end{figure}

\begin{figure}
\scriptsize{%
    \centering
    \vspace{-0.1cm}
    \begin{subfigure}[l]{.45\textwidth}
        \includegraphics[width=\textwidth]{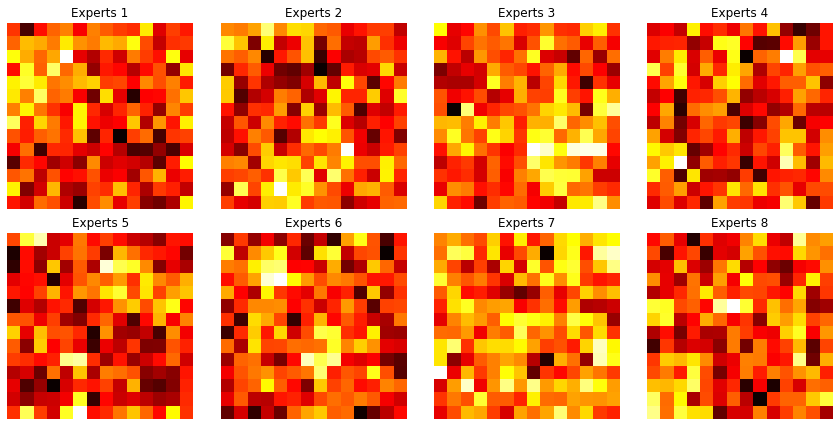}
        \subcaption{ViT-B deepest \moe{}. No absolute positional bias is observed.}\label{fig:moe_spatdistvit}
    \end{subfigure}
    \begin{subfigure}[c]{.45\textwidth}
        \includegraphics[width=\textwidth]{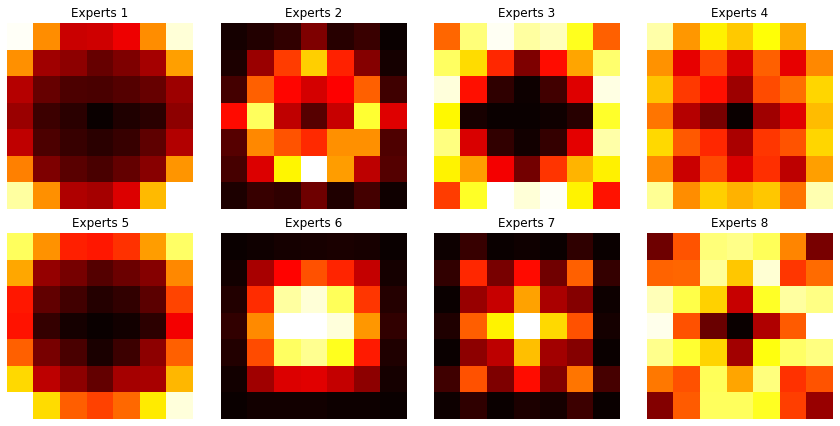}
         \subcaption{\ConvNeXt-B last \moe{} layer. A clear spatial partitioning of experts emerges.}\label{fig:moe_spatdistcvxt}
    \end{subfigure}
    \begin{subfigure}[r]{.033\textwidth}
        \centering
        \includegraphics[width=\textwidth]{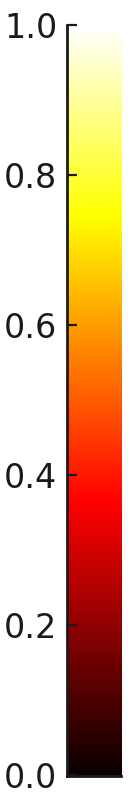}
        \hspace{10em}
    \end{subfigure}
    \caption{Average spatial distribution of experts trained on ImageNet-22K.}\label{fig:moe_spatdist}
}
\end{figure}

\textbf{Visual inspection: }\cref{fig:moe_vizu} provides a visualization of the portions of images that are assigned to the different experts. This reveals that some experts specialize in specific elements of an image, such as animals, buildings, or objects. This trend is especially apparent when considering the top-1 \moe{}, but the distinction becomes less clear with the top-2 \moe{}. In both instances, the clusters formed by each expert are challenging to interpret, and the spatial locations of each expert tend to lack clarity and can be quite ambiguous. \change{However, when directly examining the spatial distribution, a clear spatial decomposition emerges for ConvNeXt models (cf. \cref{fig:moe_spatdist}). In particular, \cref{fig:moe_spatdistcvxt} shows that certain experts focus specifically on the image borders or the center. This behavior may be explained by the strong local spatial bias imposed by convolution, in contrast to the attention-based models, which do not exhibit such patterns.}

\begin{figure}
    \centering
    \begin{subfigure}[c]{0.25\linewidth}
        \includegraphics[width=\textwidth, trim={0 0 4cm 0}, clip]{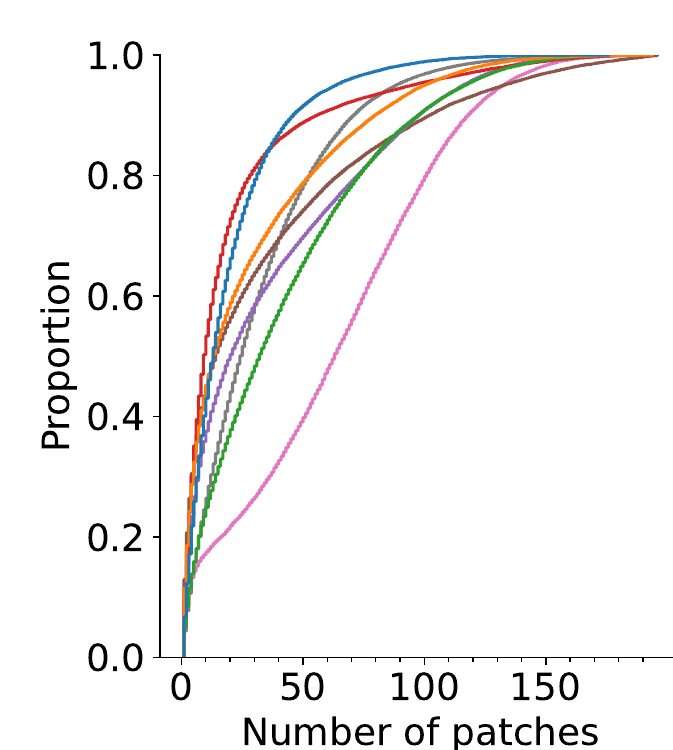}
        \subcaption{8 experts}
    \end{subfigure}
    \begin{subfigure}[c]{0.25\linewidth}
        \includegraphics[width=\textwidth, trim={0 0 4cm 0}, clip]{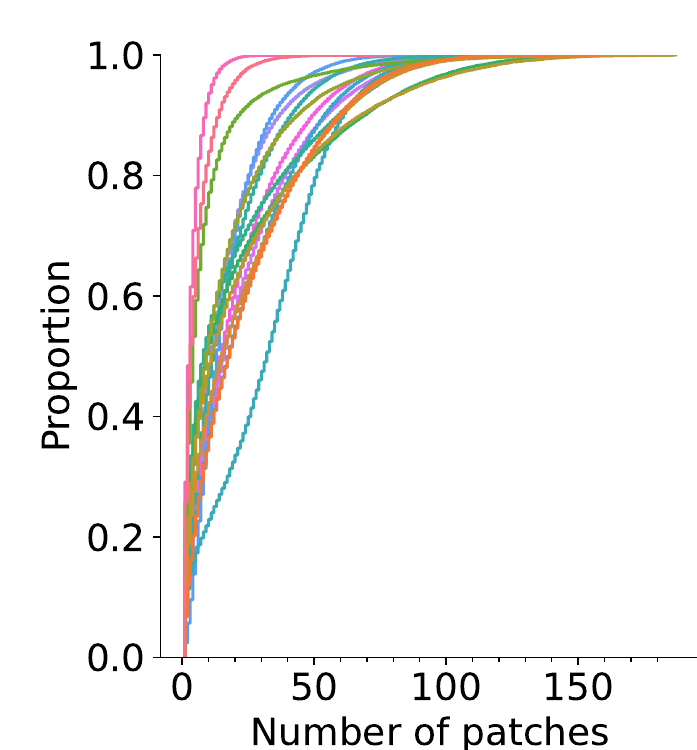}
        \subcaption{16 experts}
    \end{subfigure}
    \begin{subfigure}[c]{0.25\linewidth}
        \includegraphics[width=\textwidth, trim={0 0 4cm 0}, clip]{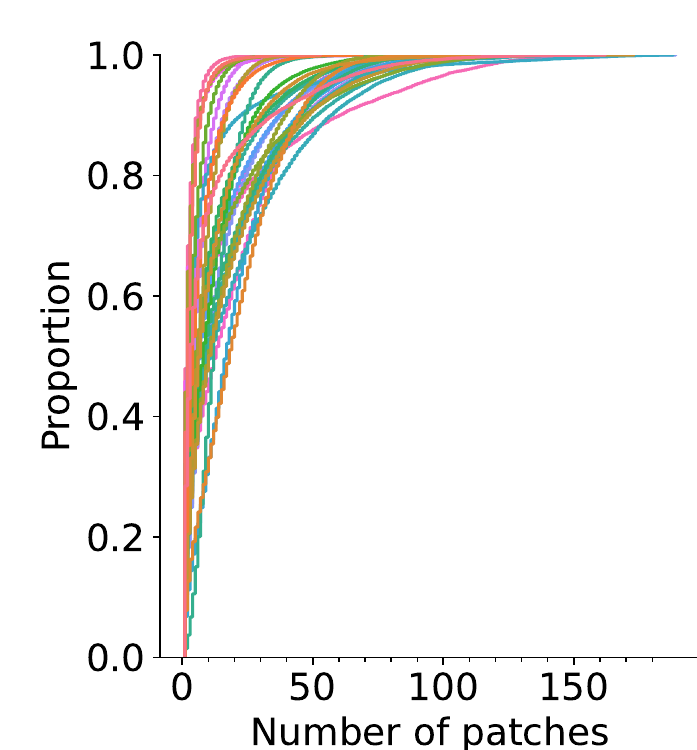}
        \subcaption{32 experts}
    \end{subfigure}
    \caption{Cumulated Distribution Function of the number of patches used per image for \ConvNeXt-S trained on \imagenettwok{}, with 8, 16, and 32 experts. Each colour corresponds to one expert. Sample detailed interpretation: for 8 experts, the ``red'' expert sees between 1 and 30 patches (out of 196) for 80\% of the images on which it is active. In particular (and even more so when there are many experts), an expert frequently covers a limited part of the image. For instance, if the y-axis is close to 1 at x-axis=100 patches, covering more than 100 patches over 196 is rare: experts don't specialize much. As the number of experts increases, this tendency of experts being used only on a few patches increases too. For 32 experts, most of them focus on less than 30 patches; 80\% of the time, they are active.}
    \label{fig:moe_cdf}
\end{figure}

\textbf{Number of experts involved per image: } We quantified the average number of experts contributing to processing an image. In the initial layers, the majority of experts are typically engaged for each image. With fewer than eight experts, most are utilized, even in deeper layers. For instance, in the ViT model, with eight experts, an average of seven experts contributed per image, while in the \ConvNeXt\ model, there are six. As we increase the number of available experts per \moe{} layer, the number of contributing experts per image also rises. For example, in \ConvNeXt{}, with 16 experts, an average of 10.5 experts are involved, and with 32 experts, this number increases to 15.8. This suggests that each expert is likely focused on a limited number of image patches. This observation is confirmed by the cumulative distribution function~(CDF) of each expert per image, as depicted in~\cref{fig:moe_cdf}. \change{On the other hand, when directly examining the routing distribution across experts, we observe a well-balanced assignment, indicating that the load-balancing loss is effectively fulfilling its role, as depicted in~\cref{fig:dist}.}

\begin{figure*}
    \centering
    \vspace{-0.5cm}
    \begin{subfigure}[c]{.16\textwidth}
        \includegraphics[width=\textwidth, trim={0 0 5.3cm 0}, clip]{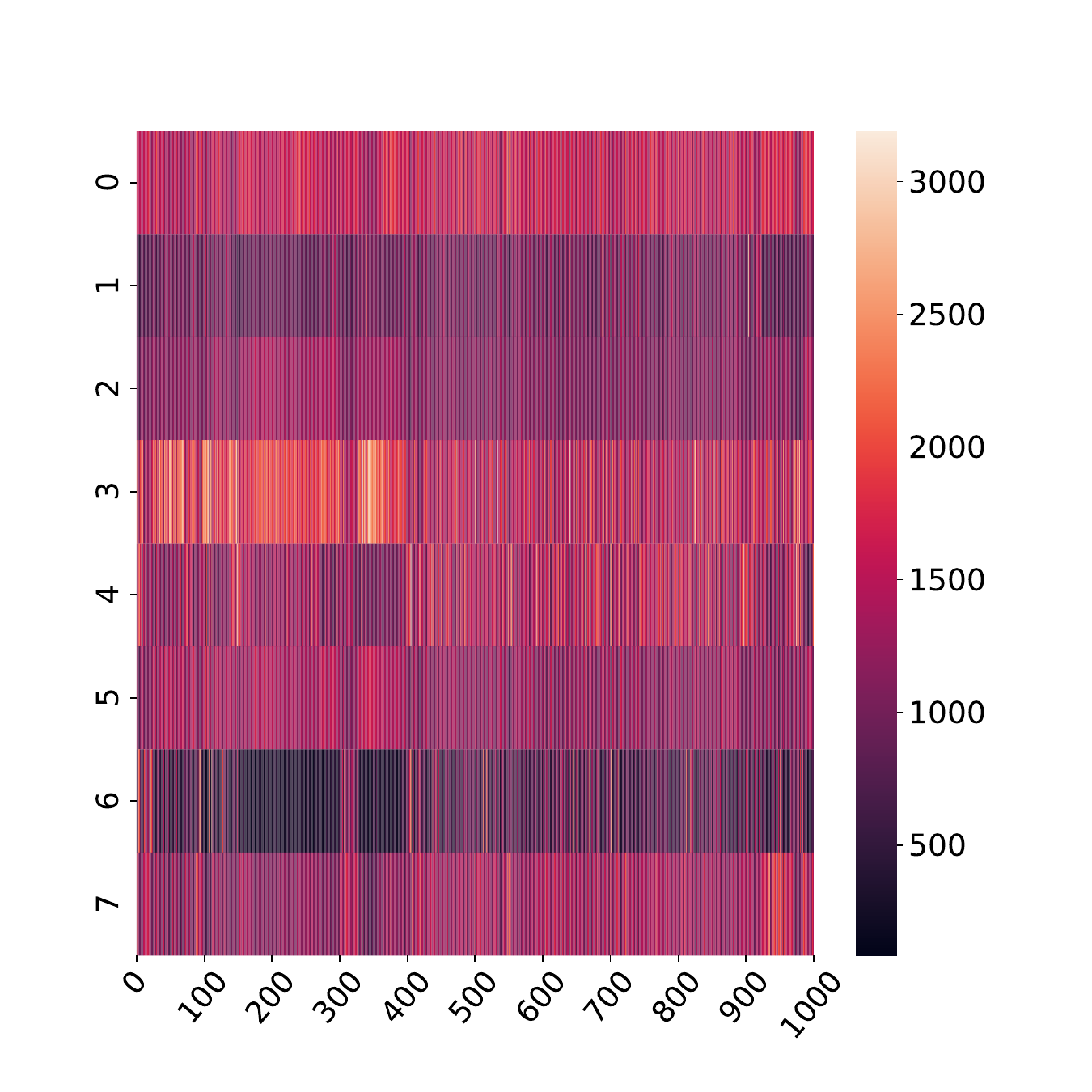}
    \end{subfigure}
    \begin{subfigure}[c]{.16\textwidth}
        \includegraphics[width=\textwidth, trim={0 0 5.3cm 0}, clip]{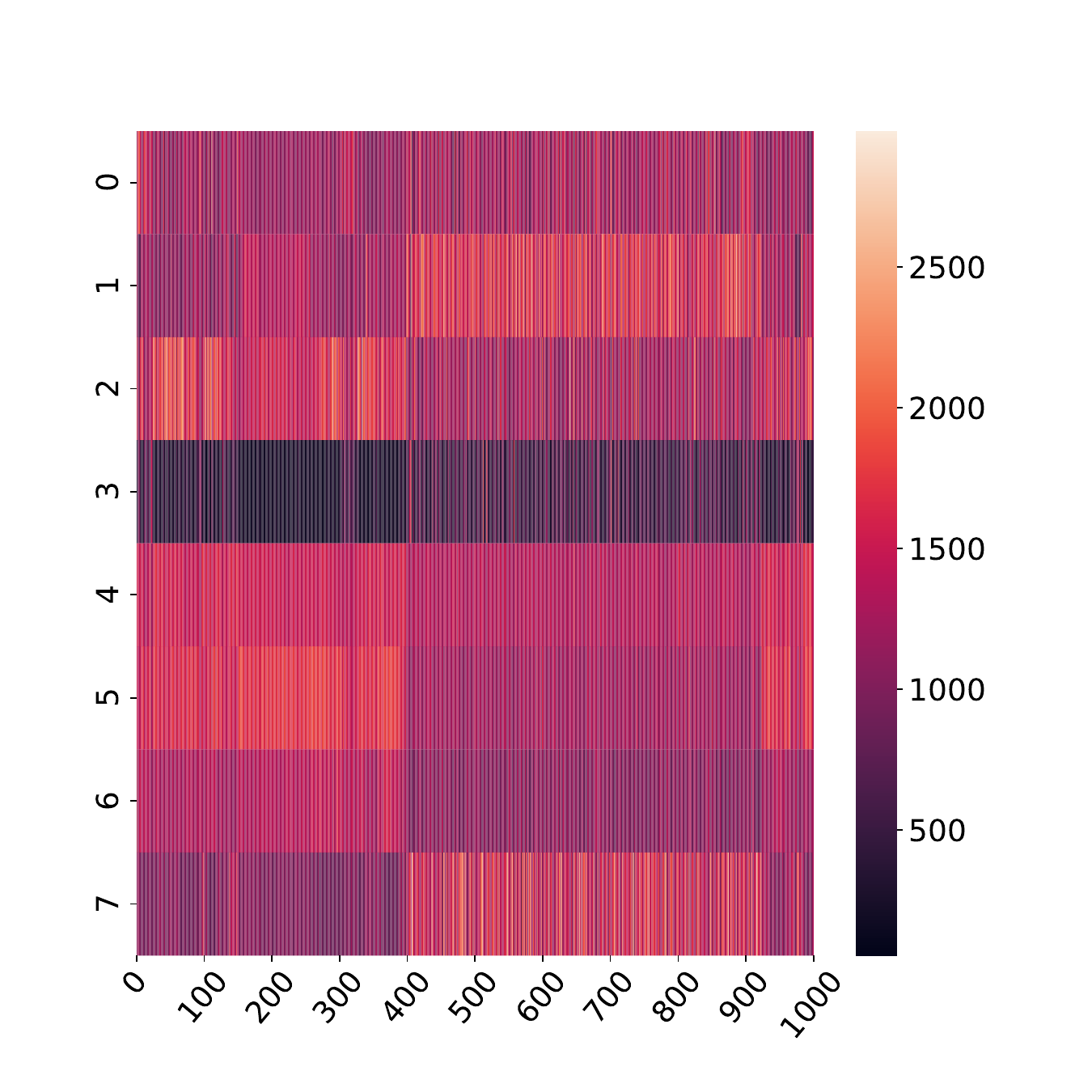}
    \end{subfigure}
    \begin{subfigure}[c]{.16\textwidth}
        \includegraphics[width=\textwidth, trim={0 0 5.3cm 0}, clip]{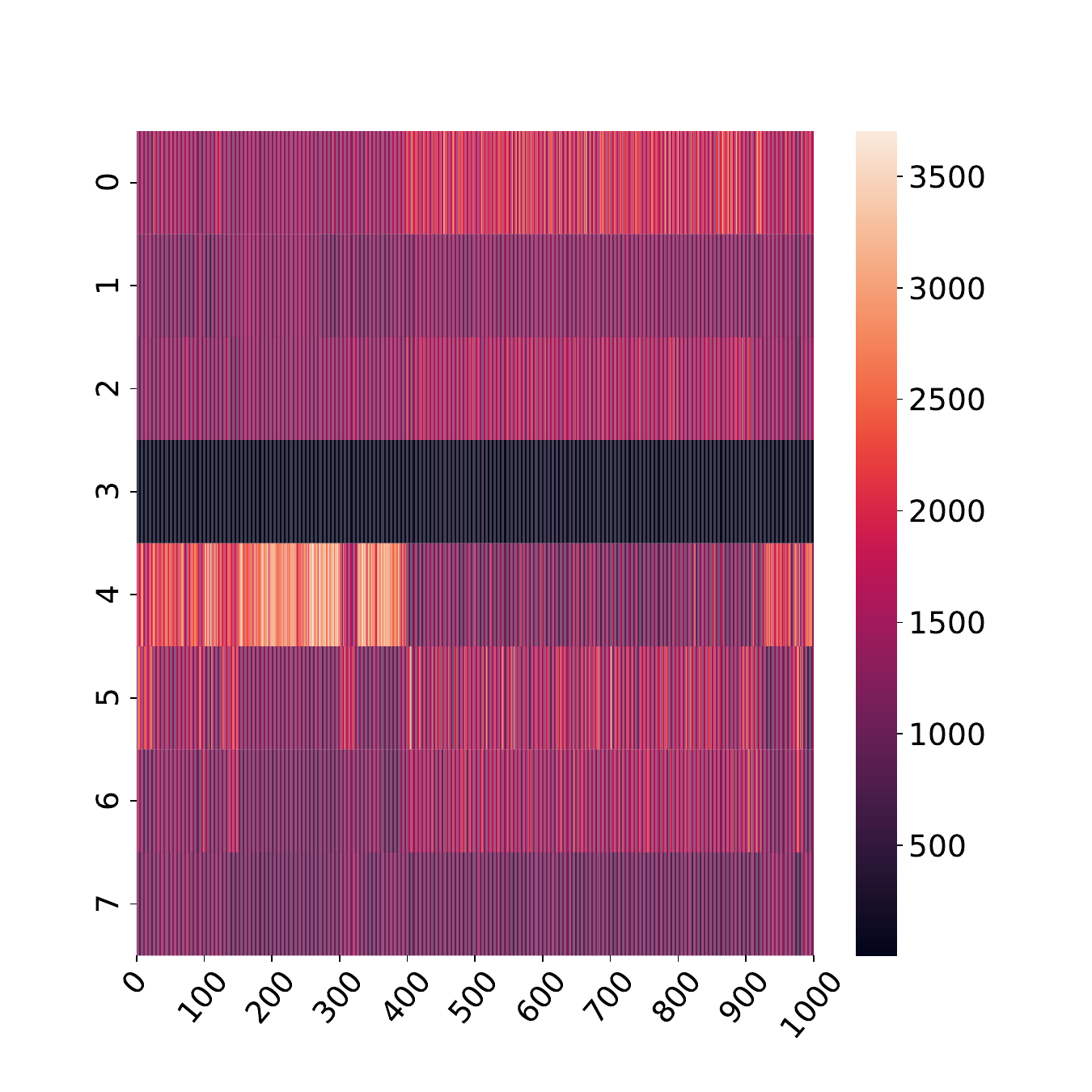}
    \end{subfigure}
    \begin{subfigure}[c]{.16\textwidth}
        \includegraphics[width=\textwidth, trim={0 0 5.3cm 0},clip]{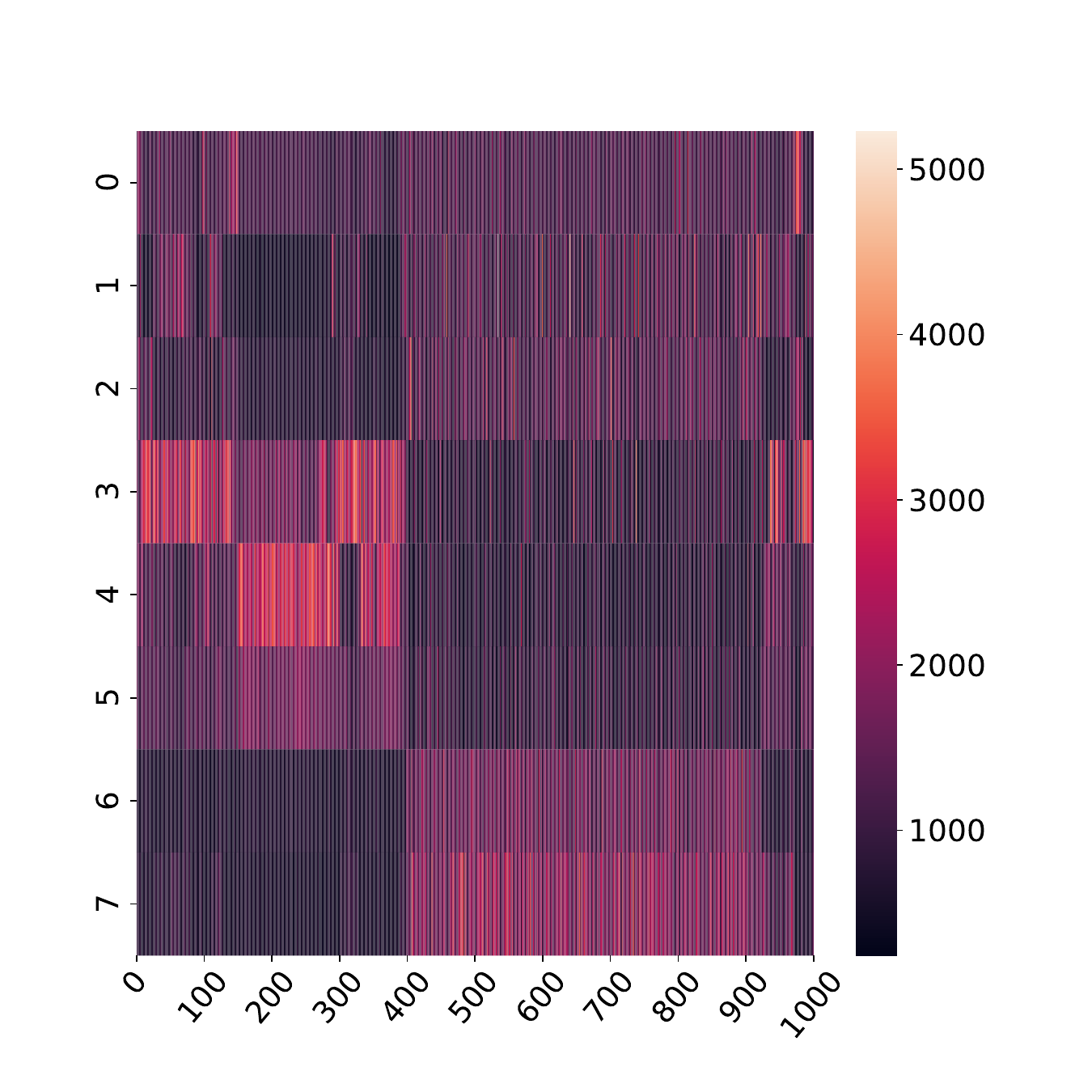}
    \end{subfigure}
    \begin{subfigure}[c]{.16\textwidth}
        \includegraphics[width=\textwidth, trim={0 0 5.3cm 0}, clip]{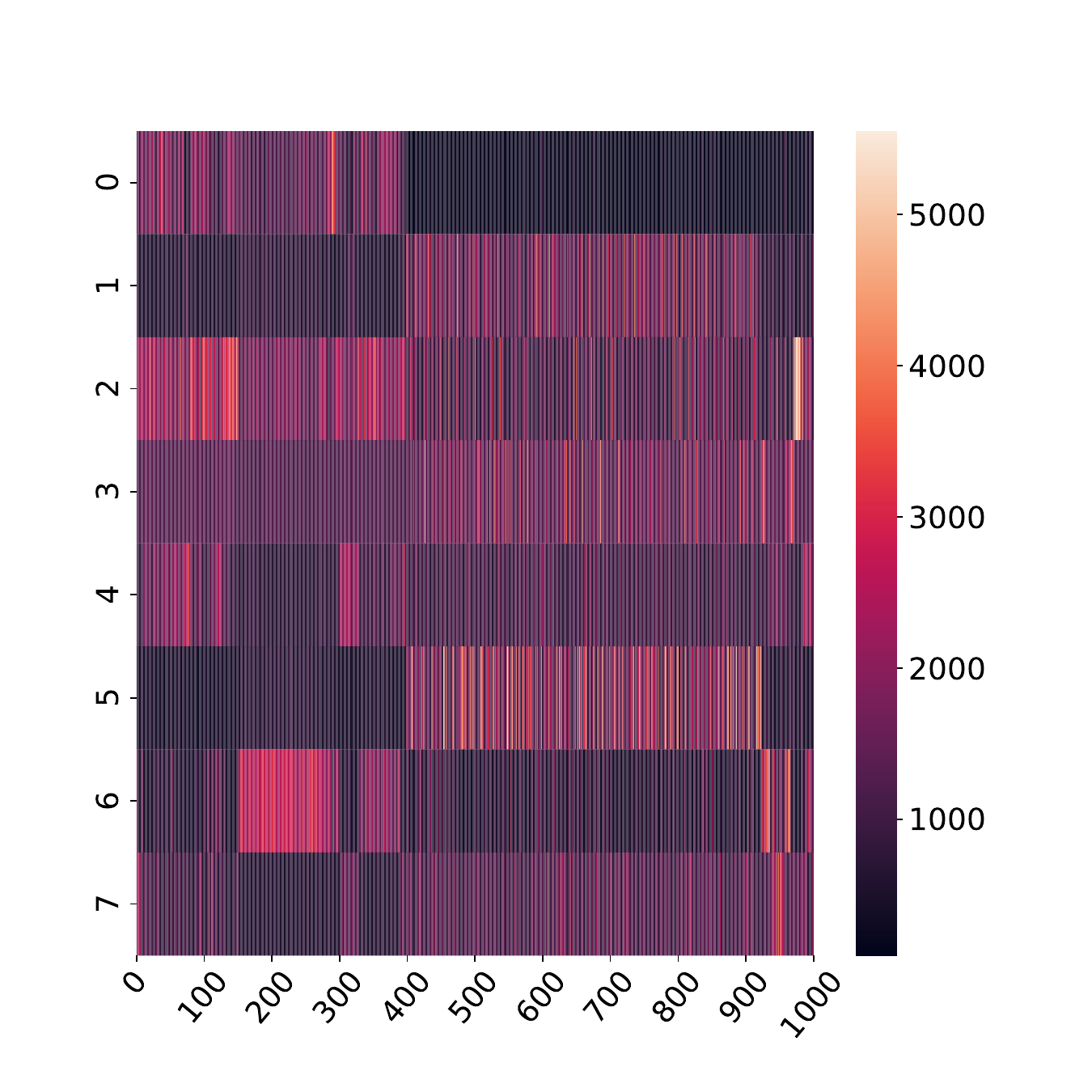}
    \end{subfigure}
    \begin{subfigure}[c]{.16\textwidth}
        \includegraphics[width=\textwidth, trim={0 0 5.3cm 0}]{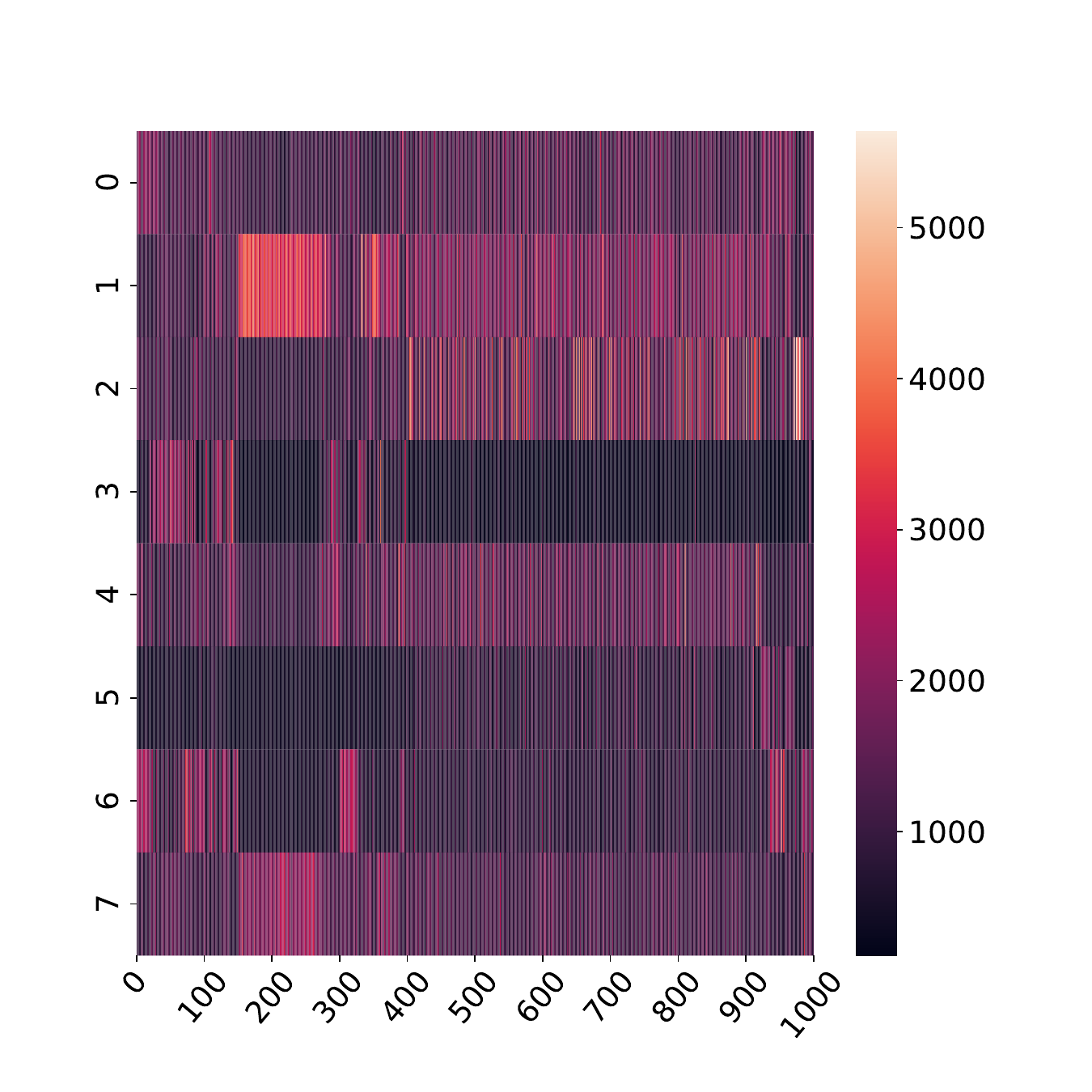}
    \end{subfigure}
    \begin{subfigure}[c]{.16\textwidth}
        \includegraphics[width=\textwidth, trim={0 0 5.3cm 0}, clip]{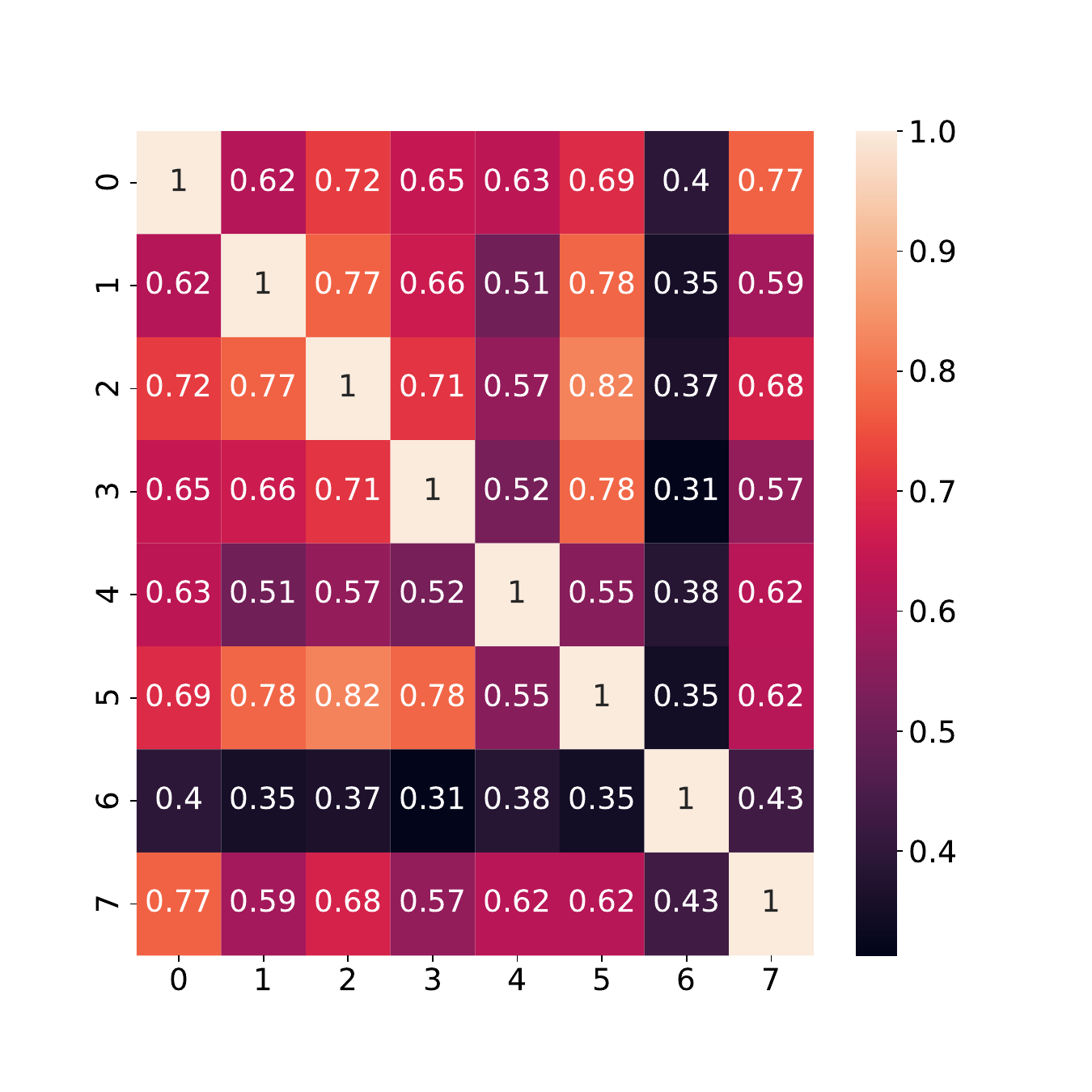}
        \subcaption{}
    \end{subfigure}
    \begin{subfigure}[c]{.16\textwidth}
        \includegraphics[width=\textwidth, trim={0 0 5.3cm 0}, clip]{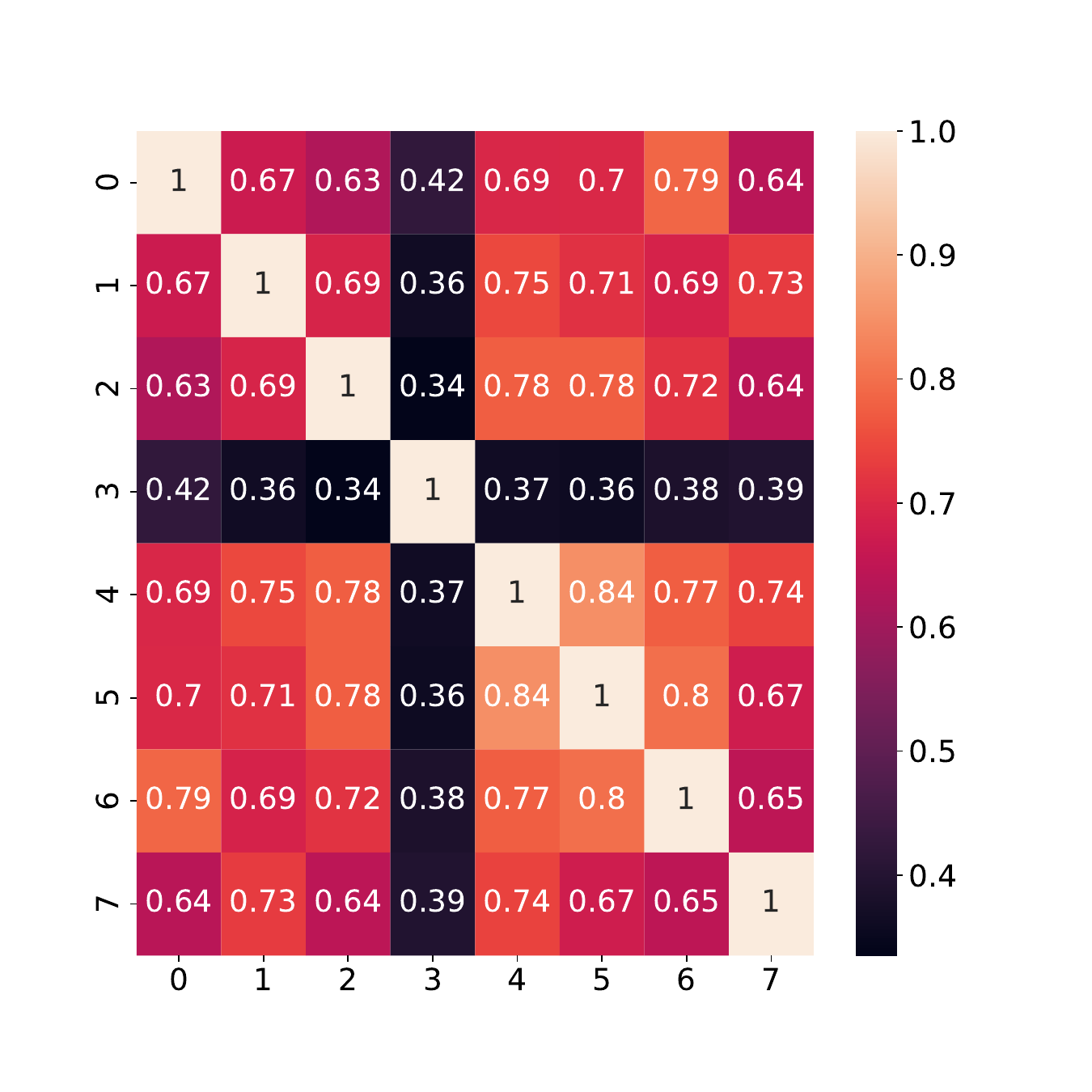}
        \subcaption{}
    \end{subfigure}
    \begin{subfigure}[c]{.16\textwidth}
        \includegraphics[width=\textwidth, trim={0 0 5.3cm 0}, clip]{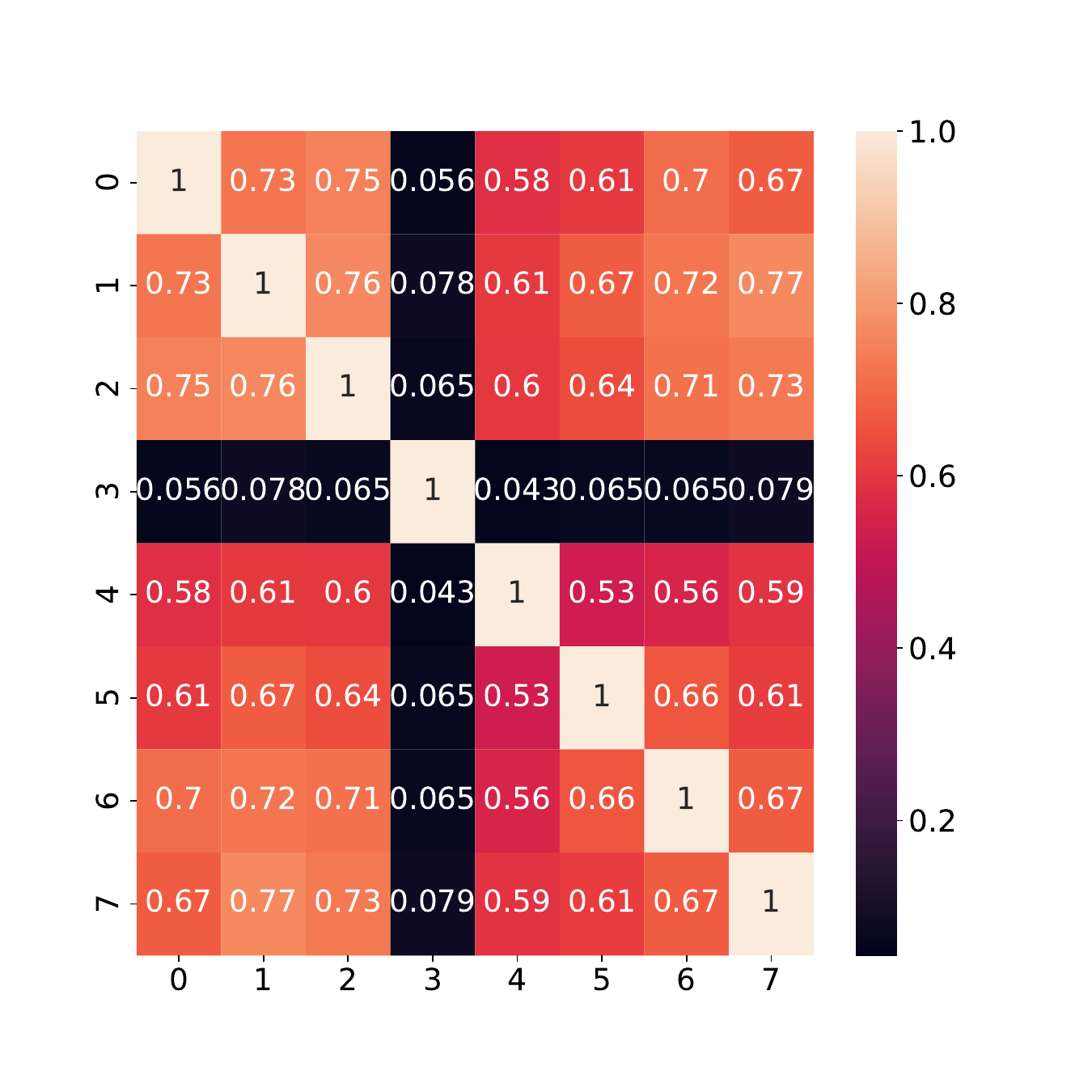}
        \subcaption{}
    \end{subfigure}
    \begin{subfigure}[c]{.16\textwidth}
        \includegraphics[width=\textwidth, trim={0 0 5.3cm 0}, clip]{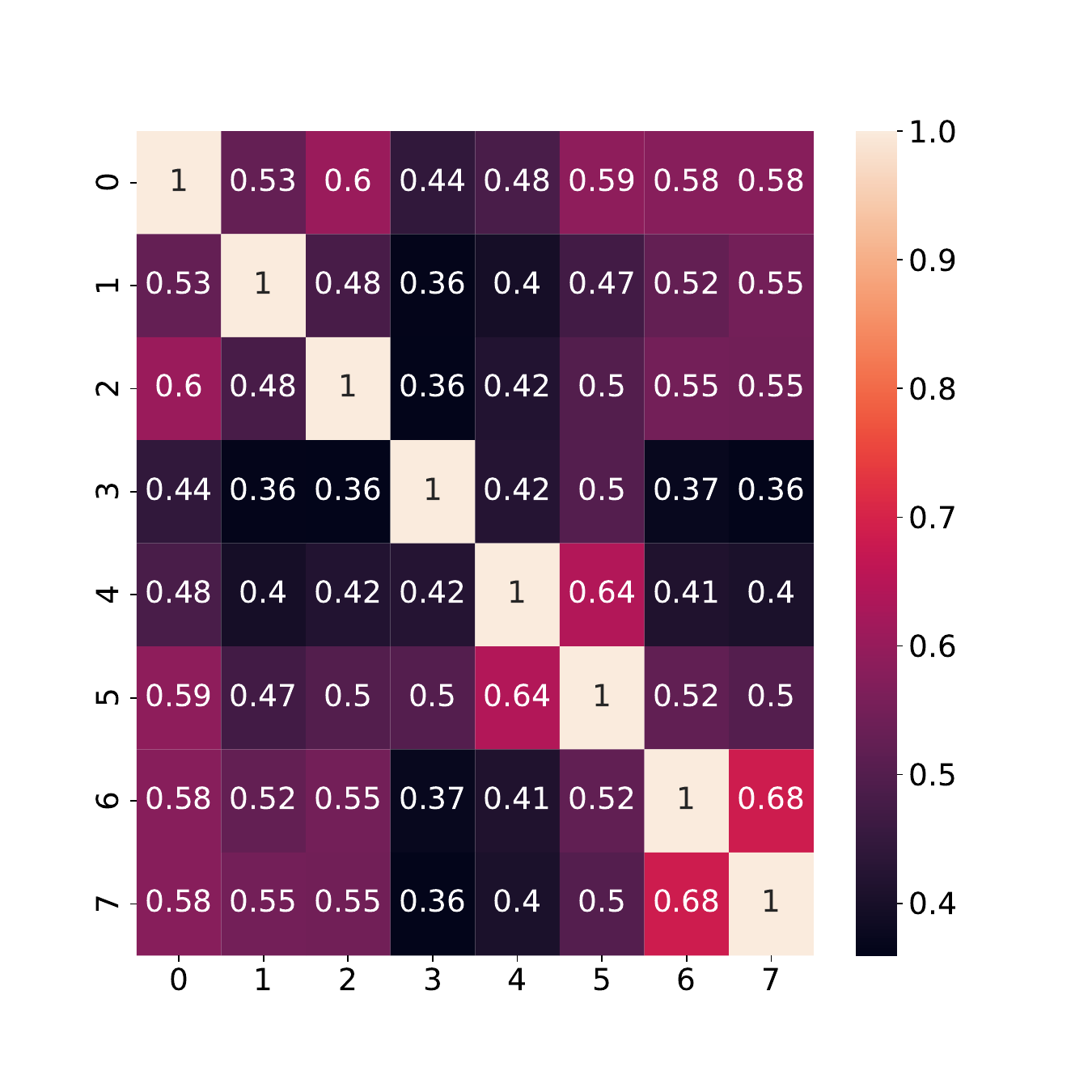}
        \subcaption{}
    \end{subfigure}
    \begin{subfigure}[c]{.16\textwidth}
        \includegraphics[width=\textwidth, trim={0 0 5.3cm 0}, clip]{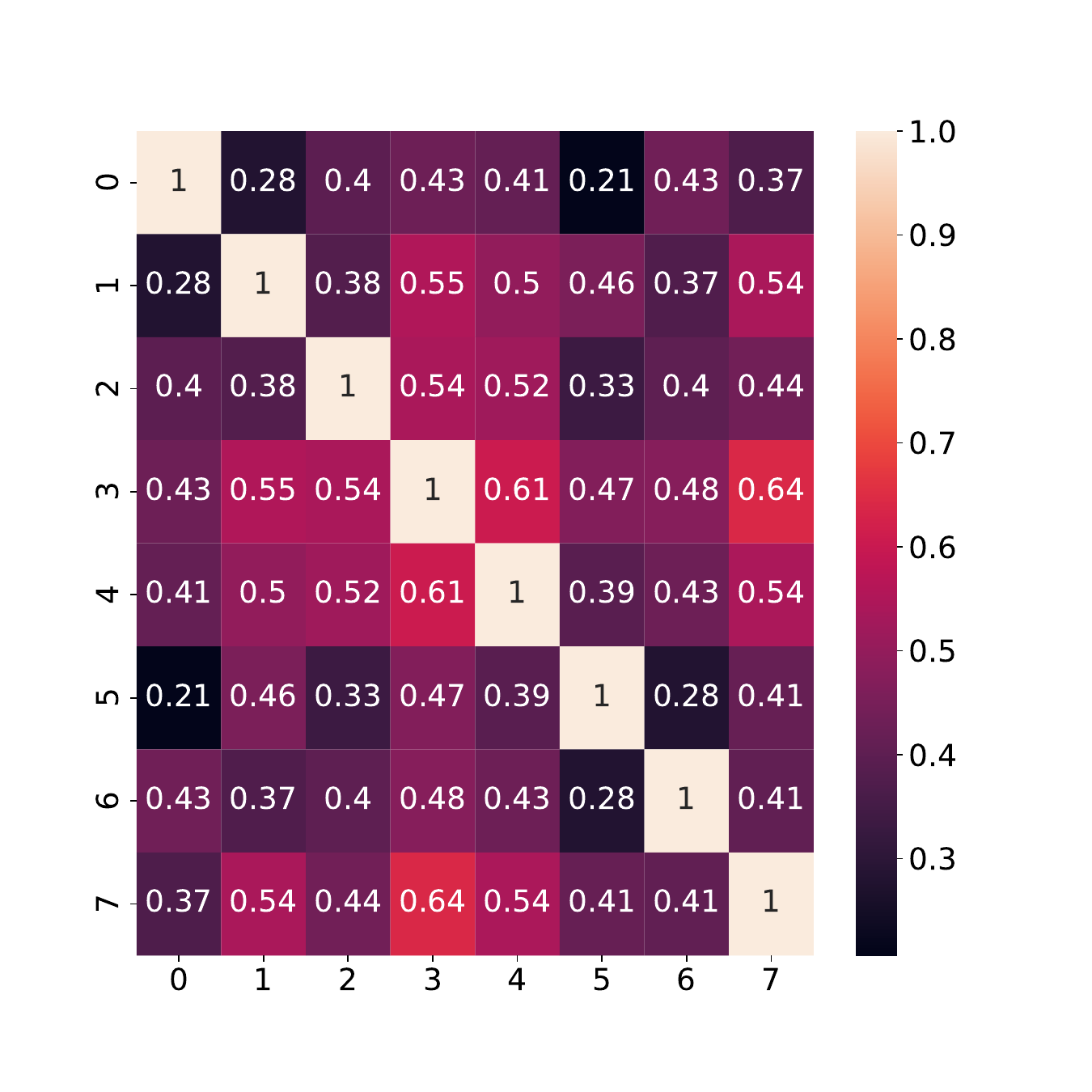}
        \subcaption{}
    \end{subfigure}
    \begin{subfigure}[c]{.16\textwidth}
        \includegraphics[width=\textwidth, trim={0 0 5.3cm 0}]{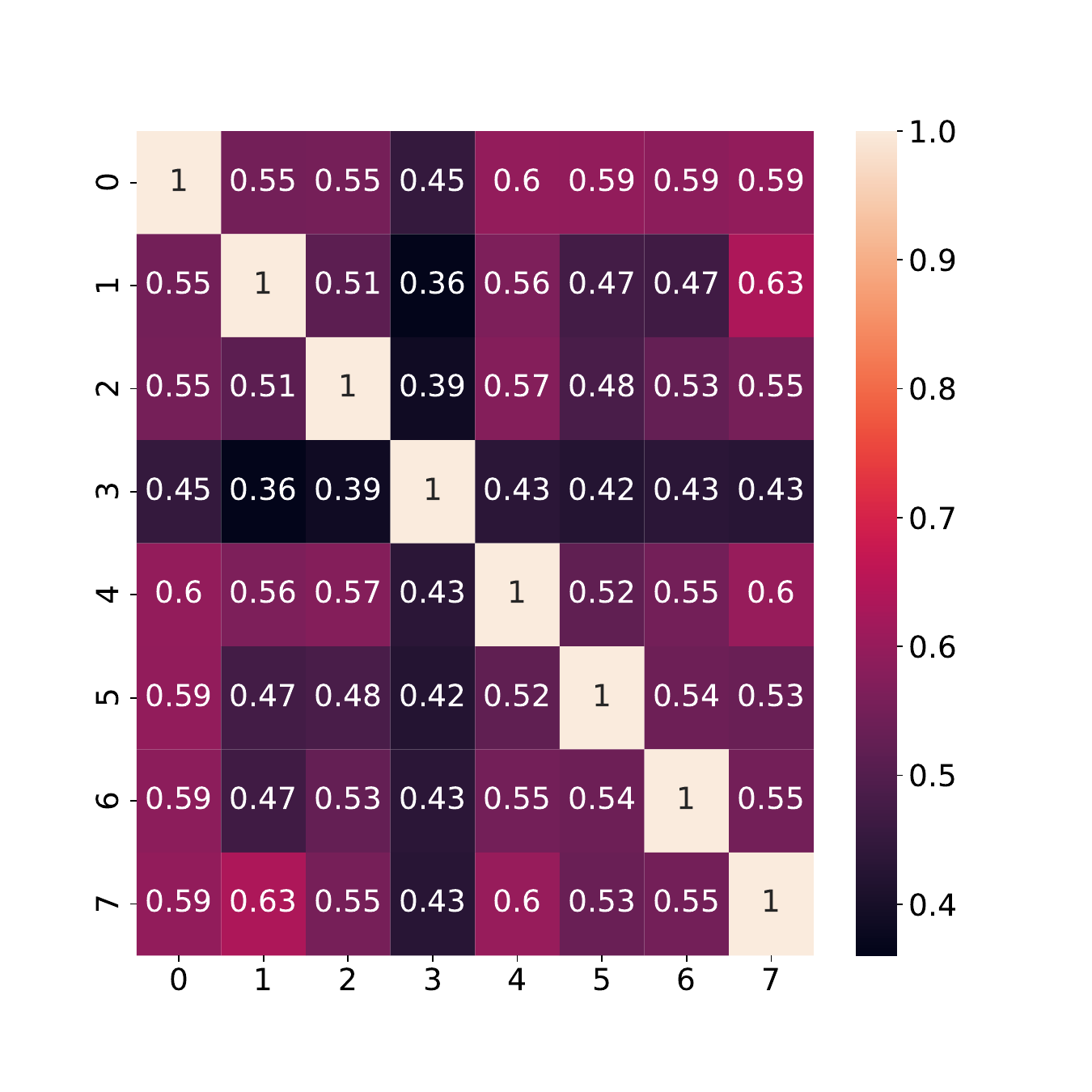}
        \subcaption{}
    \end{subfigure}
    \vspace{-0.3cm}
    \caption{Routing analysis of ViT-S-8, trained on \imagenetonek{}. Top row: per-class occurrence of the experts, with the abscissa indicating the class ID and the ordinate representing the expert index (8 experts). Bottom row: similarity score 
    between the experts. The graphs are arranged from left to right from the \moe{} layer closest to the input (a) to the layer closest to the output (f).}
    \label{fig:moe_per_class}
\end{figure*}

\begin{figure}[!htb]  
    \centering
    \begin{subfigure}[c]{.3\textwidth}
        \includegraphics[width=\textwidth, trim={0 0 5.3cm 0}, clip]{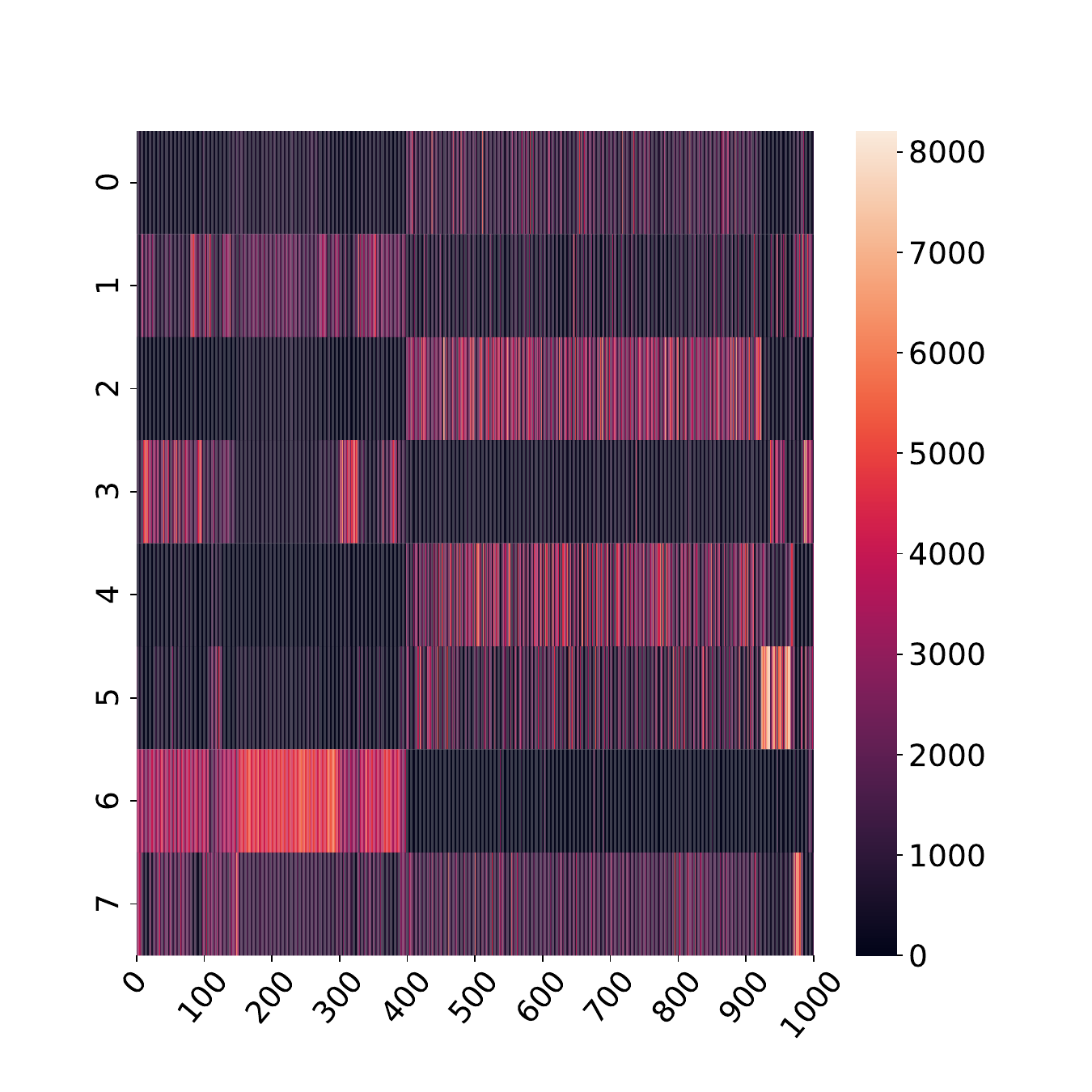}
    \end{subfigure}
    \begin{subfigure}[c]{.3\textwidth}
        \includegraphics[width=\textwidth, trim={0 0 5.3cm 0}, clip]{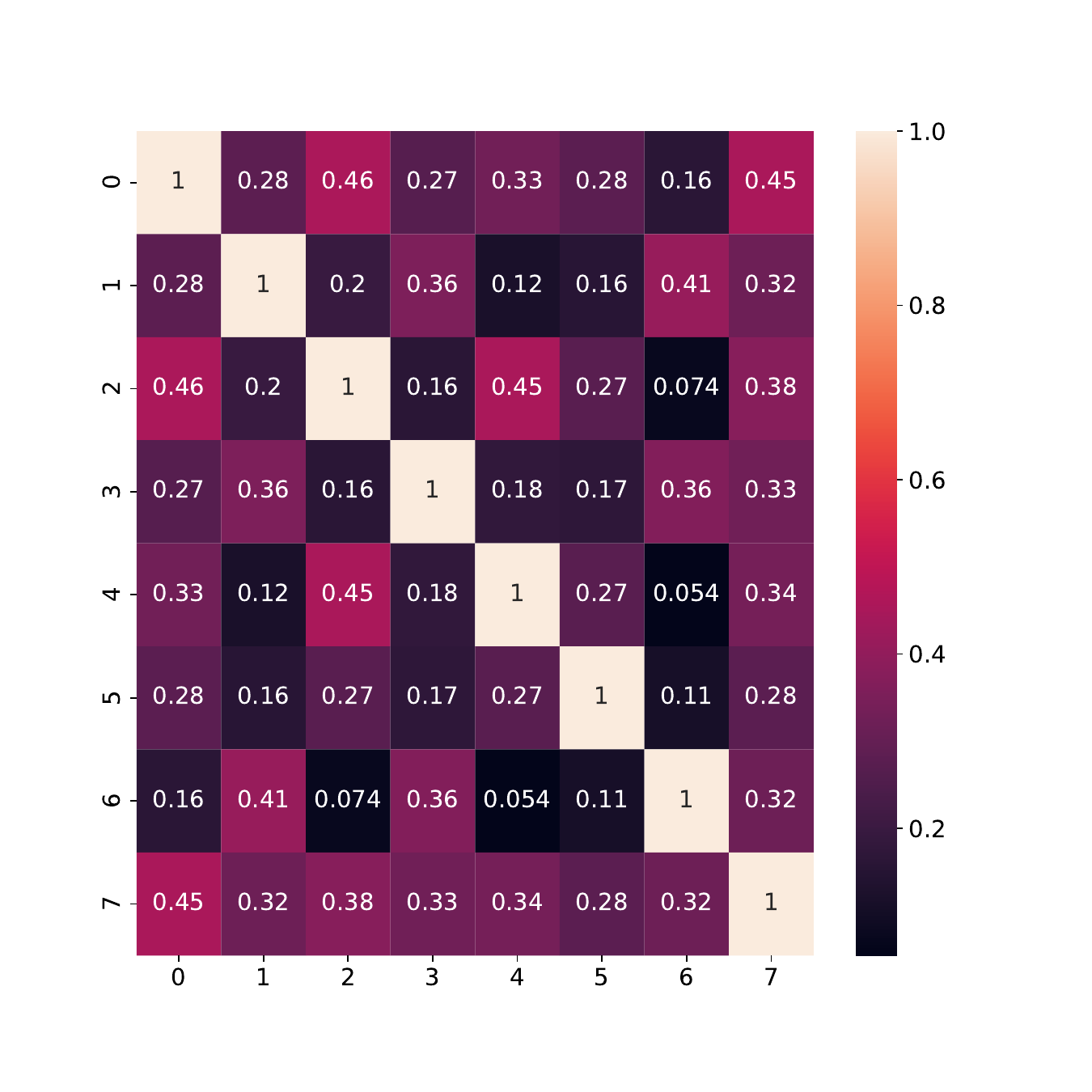}
    \end{subfigure}
    \begin{subfigure}[c]{.3\textwidth}
        \includegraphics[width=\columnwidth, trim={0 0 5.3cm 0}, clip]{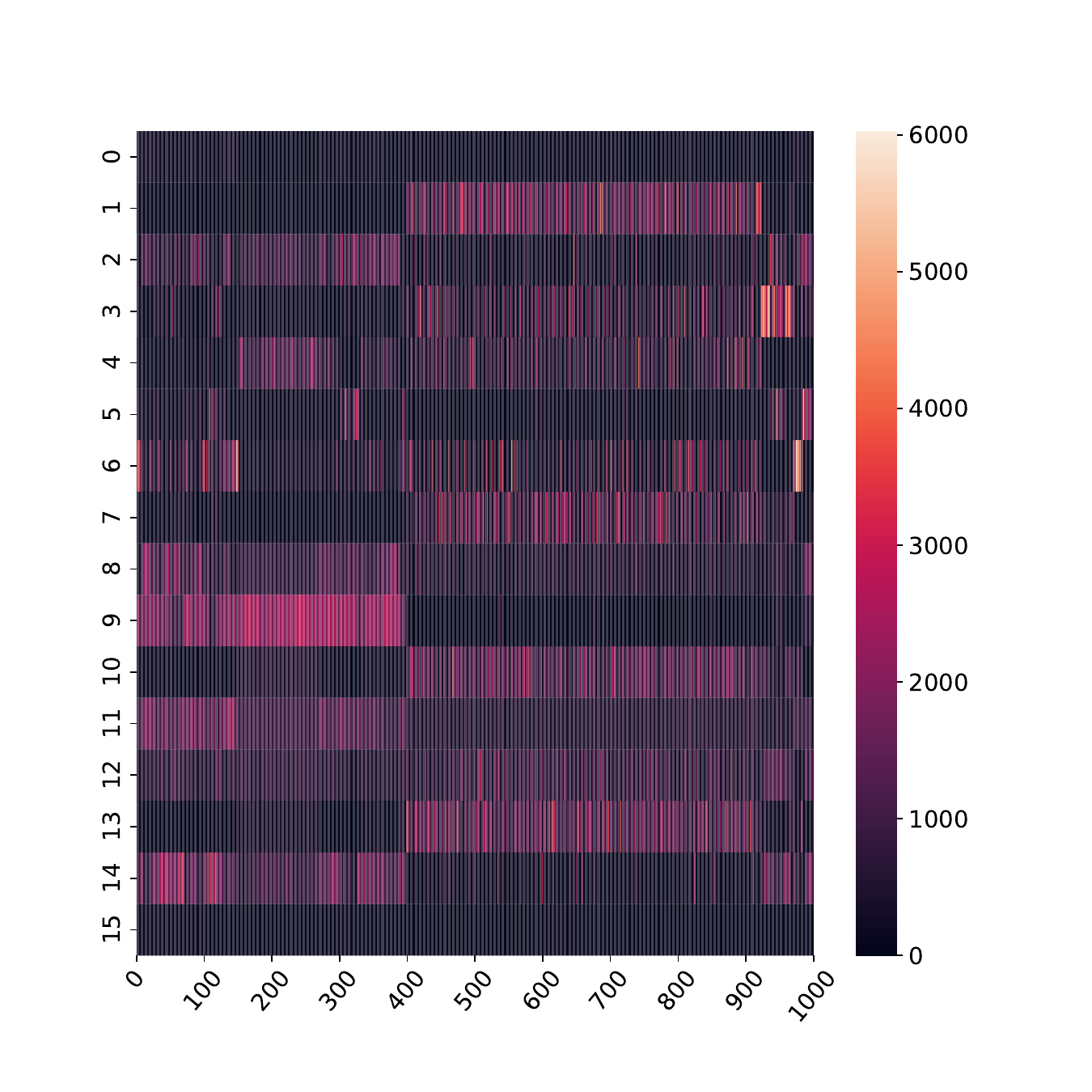}
    \end{subfigure}
    \begin{subfigure}[c]{.3\linewidth}
        \includegraphics[width=\textwidth, trim={0 0 5.3cm 0},clip]{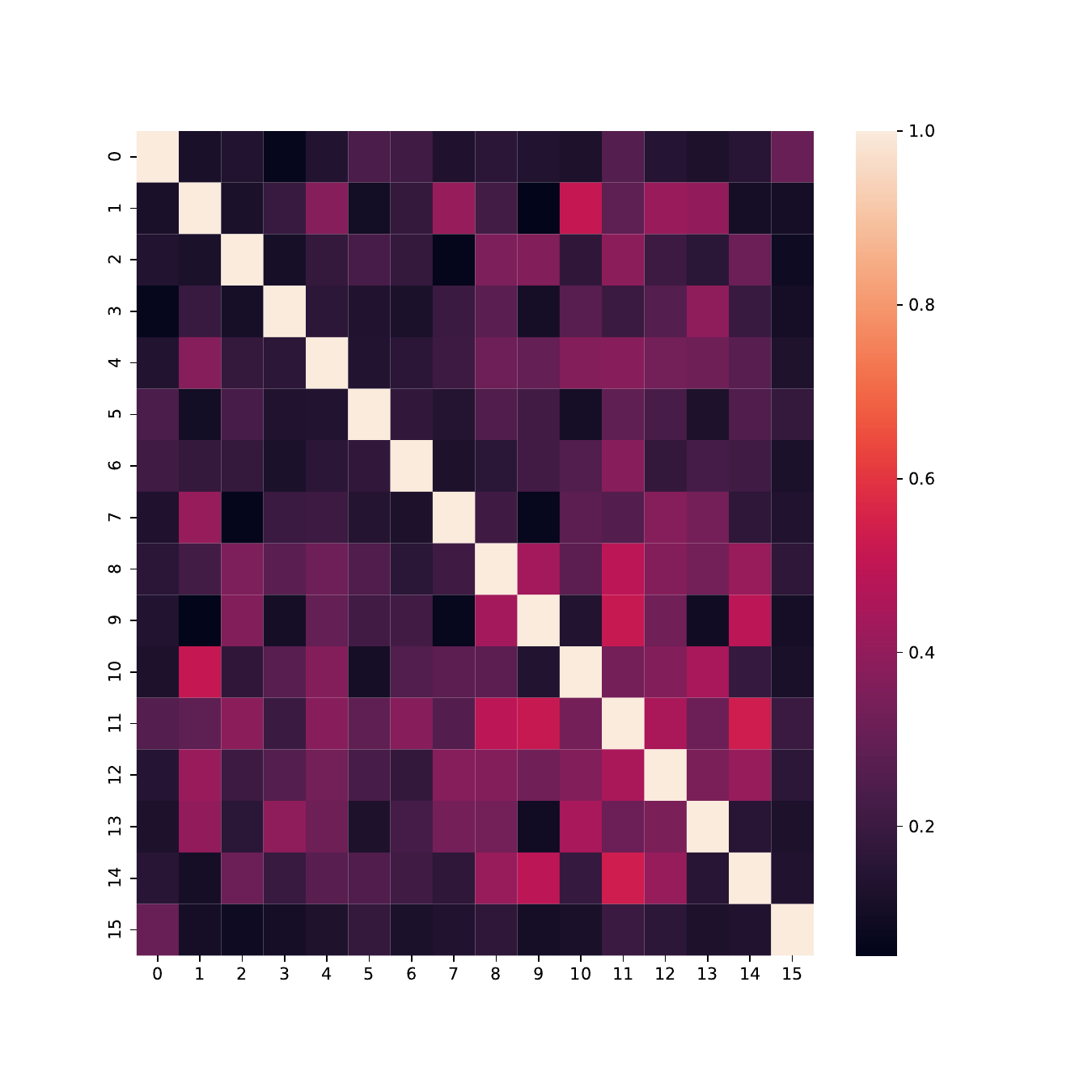}
    \end{subfigure}
    \begin{subfigure}[c]{.3\linewidth}
        \includegraphics[width=\textwidth, trim={0 0 5.3cm 0}, clip]{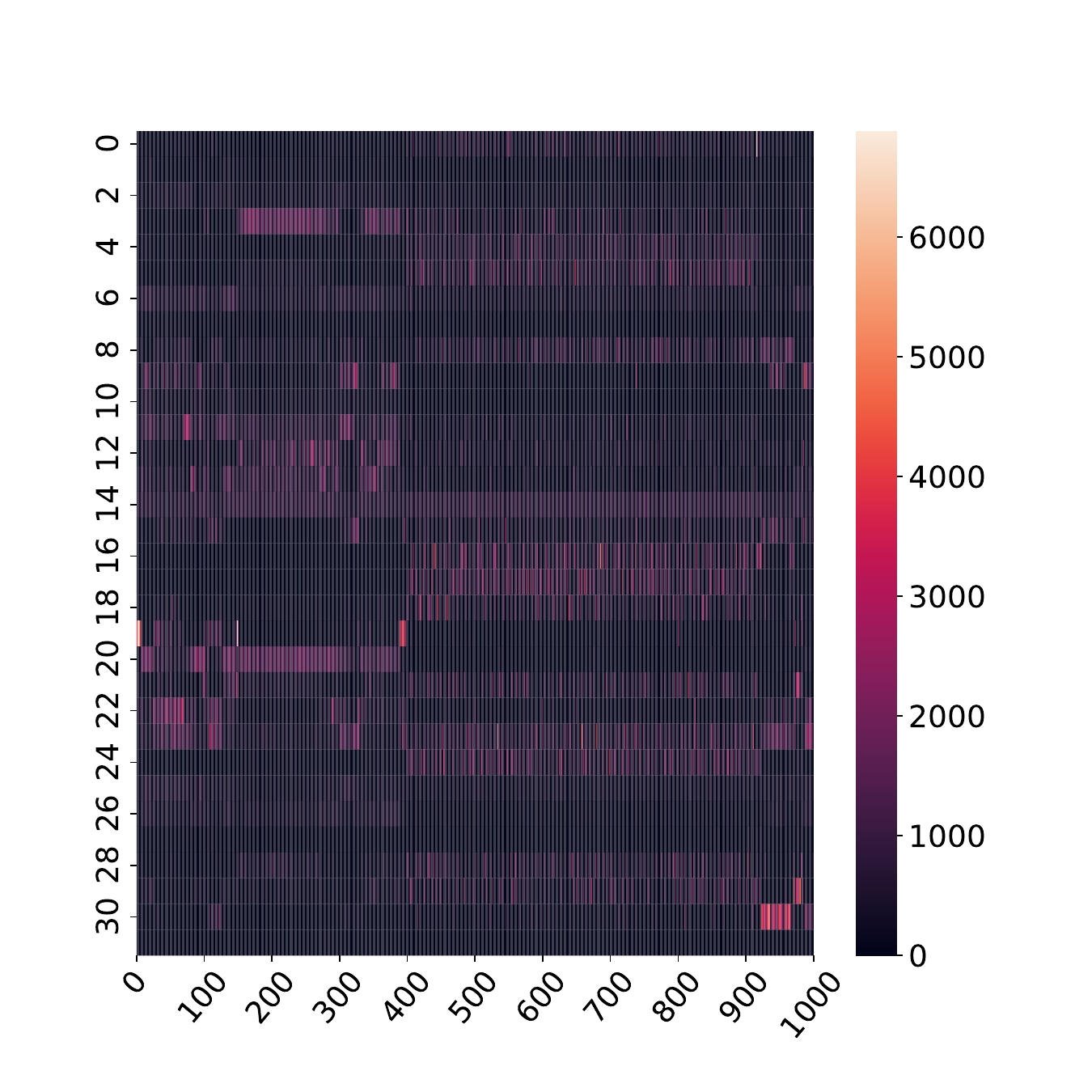}
    \end{subfigure}
    \begin{subfigure}[c]{.3\linewidth}
        \includegraphics[width=\textwidth, trim={0 0 5.3cm 0}]{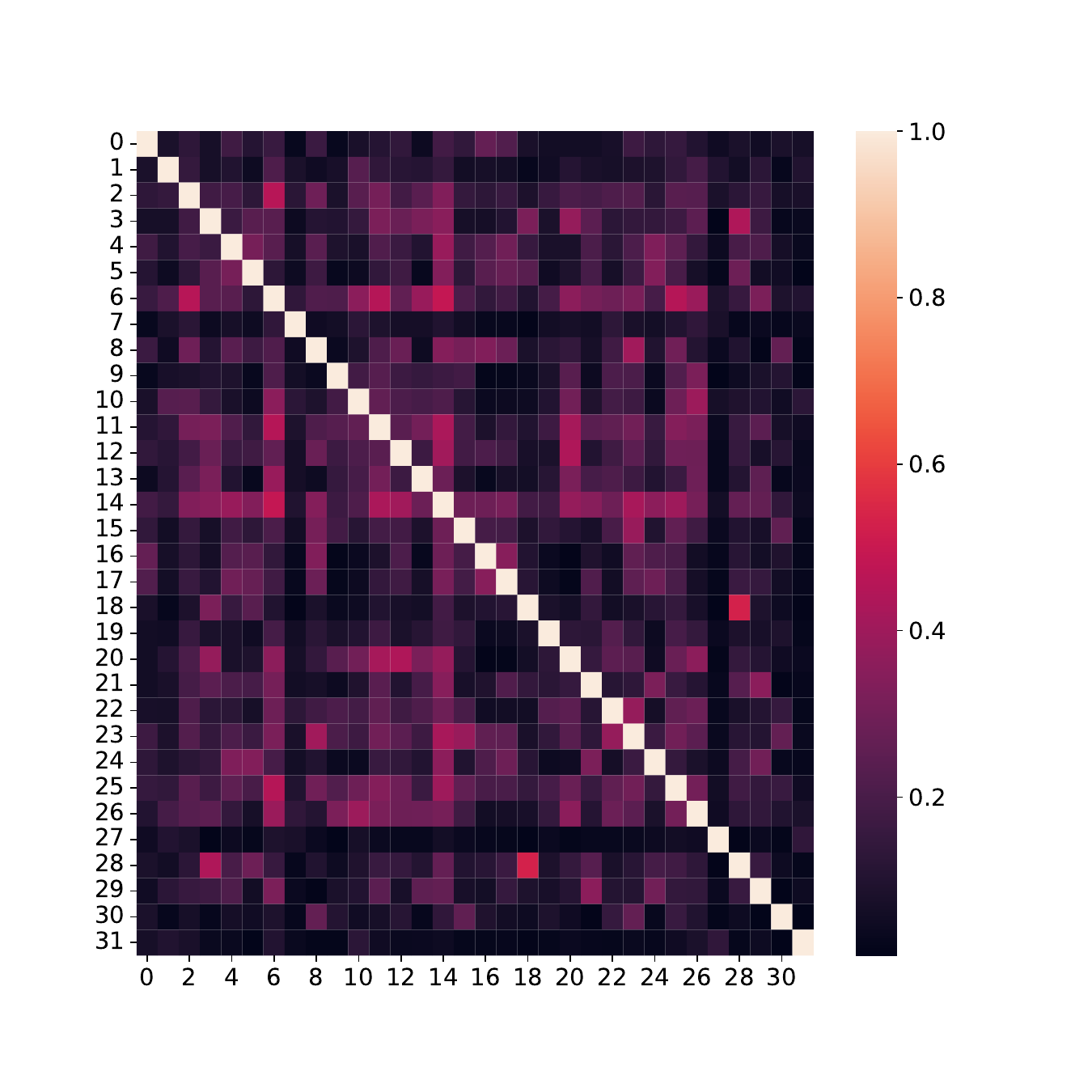}
    \end{subfigure}
    \caption{ConvNext-S trained on \imagenettwok{} with 8, 16, and 32 experts trained on \imagenettwok{} routing analysis. The right columns display the frequency of each expert across different classes, where the x-axis represents class IDs, and the y-axis denotes the expert number. The left column provides a similarity score between experts.}
    \label{fig:moe_per_class_more}
\end{figure}

\textbf{Correlation between experts and labels: }
\cref{fig:moe_per_class}~(top) presents the correlation between the number of expert occurrences and individual classes in~\imagenetonek{}. The class IDs in \imagenetonek{} are organized in such a way that adjacent classes often share similar attributes. For instance, class IDs ranging from 151 to 268 are all dedicated to different dog breeds. From~\cref{fig:moe_per_class}, we note that the occurrences in the first three \moe{} layers do not match the \imagenetonek{} classes, with most experts appearing uniformly across different classes. However, in the last three \moe{} layers, there is a noticeable trend of more experts aligning with specific \imagenetonek{} classes. This distinction is especially pronounced between experts focusing on animals (classes up to ID 397, such as experts 0 and 6 in~\cref{fig:moe_per_class}-(e)) and those centered on objects (classes beyond ID 397, like experts 1 and 5 in~\cref{fig:moe_per_class}-(e)). As the number of experts increases, this class-specific alignment still holds, though it remains challenging to clearly define the roles of these experts since they often span several scattered classes rather than forming tight, contiguous groups that would fit each class, as illustrated in~\cref{fig:moe_per_class_more}. 
\textbf{Expert Similarity: } In order to quantify the similarity between experts, we use a method derived from the Jaccard index. For each image, we count the occurrences $c_i$ of each expert $E_i$. The similarity between two experts, $E_i$ and $E_j$, is defined as \(S_{ij} = \tfrac{|E_i \cap E_j|}{|E_i \cup E_j|} = \tfrac{\sum^{E_i \cap E_j} c_i + c_j - |c_i - c_j|}{\sum^{E_i \cup E_j} c_i + c_j}\), where ${E_i \cap E_j}$ represents the images where both $E_i$ and $E_j$ appear together, while ${E_i \cup E_j}$ refers to all images where either $E_i$ or $E_j$ is present at least once. The absolute value term $|c_i-c_j|$ acts as a corrective factor to account for scenarios where one expert may be deployed for a minimal number of patches while the other is utilized for the majority, indicating a lack of synergy between the two experts. In the bottom panel of~\cref{fig:moe_per_class}, we observe a distinct pattern in the utilization of experts across the stages of the \moe{}. In the initial layers, there is a prevalent co-occurrence of multiple experts, which highlights a concentration on local image features across all the images. Conversely, as we analyze the network deeper, the frequency of expert co-occurrence diminishes. This shift indicates a transition in focus from local image elements to more semantically rich aspects, with individual experts concentrating on specific semantic attributes unique to certain images.

\textbf{Are experts well partitioned and clustered?}
%
~For a general task such as \imagenetonek{}, there is no clear, natural decomposition into experts, making it difficult to precisely assess if one expert decomposition is better than another, aside from comparing final accuracy. This comparison, however, provides limited insight into the quality of the experts. The literature~\citep{chen2023mod, mittal2022modular, mohammed2022models} suggests that, without explicit enforcement, \moe{} models struggle to separate tasks or find natural clusters beneath them. For instance, when \alessandro{a} task separation was explicitly incorporated into the loss, there was a marked improvement in both accuracy and the ease with which experts could be pruned for specific tasks~\citep{chen2023mod}. Similarly, \citet{mittal2022modular} found that the enforced decomposition resulted in substantial gains, while the absence of such enforcement left the model unable to discover optimal partitions.
Consequently, we make three observations on ImageNet. 
%
First, as the number of experts increases, there is a corresponding rise in the number of experts actively involved. Even with 32 experts, almost half of the experts are used per image in the last \moe{} layers. This seems counterintuitive since ImageNet is centered around objects; at least for deep layers, it should plateau at some point. Instead, experts are not well aligned with classes and spread widely, as depicted in~\cref{fig:moe_per_class}. 
%
Second, an analysis of the cumulative distribution function~(CDF) of experts' engagement~(\cref{fig:moe_cdf}) reveals a prevalent trend in which most experts are assigned to less than 20 patches of an image, equating to less than 10\% of the total image area, 50\% of the time. Moreover, these patches are not necessarily contiguous. 
%
Third, when we look at the visualization of each expert, like in~\cref{fig:moe_vizu}, we observe a concentration of experts in the image for \moe{} layers near the output. The interpretation of each expert's role becomes challenging: An expert does not seem to adhere closely to an object, or to a part of the location of an object.
\begin{AIbox}{Lack of Coherent Expert Partitioning}
    Qualitative and quantitative analyses reveal no consistent expert partitioning in vision \moe{}, particularly as the number of experts increases. With few experts, routing weakly aligns with ImageNet classes, but as the scale grows, expert allocation becomes diffuse, many experts activate per image, each covering only a small portion of patches, and long-range consistency across class IDs becomes less clear. \ConvNeXt\ shows some spatial biases, whereas ViT shows none; in both, assignments remain well load-balanced yet not semantically disentangled, suggesting that the observed gains stem from added capacity rather than interpretable specialization.
\end{AIbox}

\section{Conclusion}
%
Implementing effective~\MixtureOfExperts{}~(\moe{}) models for image classification tasks remains a challenging and open problem. Our experiments show that, particularly in large-scale models, \moe{} for image classification offers limited gains over state-of-the-art methods in accuracy, robustness, generalization, or per-sample efficiency across both convolutional networks and vision transformers.
%
%
This contrasts with NLP tasks, where \moe{} improves performance by enhancing capacity for knowledge storage and retrieval. In image classification, which relies more on feature extraction and extrapolation, \moe{} models show only modest benefits, mainly in smaller models with fewer activated parameters. These findings suggest that the effectiveness of \moe{} in image classification tasks is highly context- and scale-dependent, with advantages diminishing as model capacity grows.
%

\subsubsection*{Acknowledgements}
The authors thank the reviewers for their insightful and constructive feedback, which has substantially improved the quality of this work. AL acknowledges support from the French National Research Agency (ANR) under Grant No. ANR-23-CPJ1-0099-01.
\FloatBarrier
\bibliographystyle{tmlr}
\bibliography{references}

\begin{thebibliography}{42}
\providecommand{\natexlab}[1]{#1}
\providecommand{\url}[1]{\texttt{#1}}
\expandafter\ifx\csname urlstyle\endcsname\relax
  \providecommand{\doi}[1]{doi: #1}\else
  \providecommand{\doi}{doi: \begingroup \urlstyle{rm}\Url}\fi

\bibitem[Alabdulmohsin et~al.(2023)Alabdulmohsin, Zhai, Kolesnikov, and
  Beyer]{alabdulmohsin2023getting}
Ibrahim Alabdulmohsin, Xiaohua Zhai, Alexander Kolesnikov, and Lucas Beyer.
\newblock Getting {ViT} in shape: Scaling laws for compute-optimal model
  design.
\newblock \emph{arXiv:2305.13035}, 2023.

\bibitem[Chen et~al.(2023)Chen, Shen, Ding, Chen, Zhao, Learned-Miller, and
  Gan]{chen2023mod}
Zitian Chen, Yikang Shen, Mingyu Ding, Zhenfang Chen, Hengshuang Zhao, Erik~G
  Learned-Miller, and Chuang Gan.
\newblock Mod-squad: Designing mixtures of experts as modular multi-task
  learners.
\newblock In \emph{IEEE/CVF Conference on Computer Vision and Pattern
  Recognition}, pp.\  11828--11837, 2023.

\bibitem[Chen et~al.(2022)Chen, Deng, Wu, Gu, and Li]{chen2022towards}
Zixiang Chen, Yihe Deng, Yue Wu, Quanquan Gu, and Yuanzhi Li.
\newblock Towards understanding mixture of experts in deep learning.
\newblock \emph{arXiv:2208.02813}, 2022.

\bibitem[Chi et~al.(2022)Chi, Dong, Huang, Dai, Ma, Patra, Singhal, Bajaj,
  Song, Mao, et~al.]{chi2022representation}
Zewen Chi, Li~Dong, Shaohan Huang, Damai Dai, Shuming Ma, Barun Patra, Saksham
  Singhal, Payal Bajaj, Xia Song, Xian-Ling Mao, et~al.
\newblock On the representation collapse of sparse mixture of experts.
\newblock \emph{Advances in Neural Information Processing Systems},
  35:\penalty0 34600--34613, 2022.

\bibitem[Costa-juss{\`a} et~al.(2022)Costa-juss{\`a}, Cross, {\c{C}}elebi,
  Elbayad, Heafield, Heffernan, Kalbassi, Lam, Licht, Maillard,
  et~al.]{costa2022no}
Marta~R Costa-juss{\`a}, James Cross, Onur {\c{C}}elebi, Maha Elbayad, Kenneth
  Heafield, Kevin Heffernan, Elahe Kalbassi, Janice Lam, Daniel Licht, Jean
  Maillard, et~al.
\newblock No language left behind: Scaling human-centered machine translation.
\newblock \emph{arXiv:2207.04672}, 2022.

\bibitem[Cubuk et~al.(2020)Cubuk, Zoph, Shlens, and Le]{cubuk:20randaugment}
Ekin~D Cubuk, Barret Zoph, Jonathon Shlens, and Quoc~V Le.
\newblock Randaugment: Practical automated data augmentation with a reduced
  search space.
\newblock In \emph{IEEE/CVF Conference on Computer Vision and Pattern
  Recognition Workshops}, pp.\  702--703, 2020.

\bibitem[Dosovitskiy et~al.(2020)Dosovitskiy, Beyer, Kolesnikov, Weissenborn,
  Zhai, Unterthiner, Dehghani, Minderer, Heigold, Gelly, Uszkoreit, and
  Houlsby]{dosovitskiy:20}
A.~Dosovitskiy, L.~Beyer, Alexander Kolesnikov, Dirk Weissenborn, Xiaohua Zhai,
  Thomas Unterthiner, M.~Dehghani, Matthias Minderer, G.~Heigold, S.~Gelly,
  Jakob Uszkoreit, and N.~Houlsby.
\newblock An image is worth 16x16 words: Transformers for image recognition at
  scale.
\newblock \emph{International Conference on Learning Representations}, 2020.

\bibitem[Fedus et~al.(2022)Fedus, Zoph, and Shazeer]{fedus2022switch}
William Fedus, Barret Zoph, and Noam Shazeer.
\newblock Switch transformers: Scaling to trillion parameter models with simple
  and efficient sparsity.
\newblock \emph{The Journal of Machine Learning Research}, 23\penalty0
  (1):\penalty0 5232--5270, 2022.

\bibitem[Han et~al.(2024)Han, Wei, Dou, Wang, Qiang, He, Sun, Han, and
  Tian]{han2024vimoe}
Xumeng Han, Longhui Wei, Zhiyang Dou, Zipeng Wang, Chenhui Qiang, Xin He,
  Yingfei Sun, Zhenjun Han, and Qi~Tian.
\newblock Vimoe: An empirical study of designing vision mixture-of-experts.
\newblock \emph{arXiv preprint arXiv:2410.15732}, 2024.

\bibitem[Hendrycks et~al.(2021{\natexlab{a}})Hendrycks, Basart, Mu, Kadavath,
  Wang, Dorundo, Desai, Zhu, Parajuli, Guo, et~al.]{hendrycks2021many}
Dan Hendrycks, Steven Basart, Norman Mu, Saurav Kadavath, Frank Wang, Evan
  Dorundo, Rahul Desai, Tyler Zhu, Samyak Parajuli, Mike Guo, et~al.
\newblock The many faces of robustness: A critical analysis of
  out-of-distribution generalization.
\newblock In \emph{IEEE/CVF International Conference on Computer Vision}, pp.\
  8340--8349, 2021{\natexlab{a}}.

\bibitem[Hendrycks et~al.(2021{\natexlab{b}})Hendrycks, Zhao, Basart,
  Steinhardt, and Song]{hendrycks2021nae}
Dan Hendrycks, Kevin Zhao, Steven Basart, Jacob Steinhardt, and Dawn Song.
\newblock Natural adversarial examples.
\newblock In \emph{IEEE/CVF Conference on Computer Vision and Pattern
  Recognition}, pp.\  15262--15271, 2021{\natexlab{b}}.

\bibitem[Hihn \& Braun(2021)Hihn and Braun]{hihn2021mixture}
Heinke Hihn and Daniel~A Braun.
\newblock Mixture-of-variational-experts for continual learning.
\newblock \emph{arXiv:2110.12667}, 2021.

\bibitem[Hwang et~al.(2023{\natexlab{a}})Hwang, Cui, Xiong, Yang, Liu, Hu,
  Wang, Salas, Jose, Ram, Chau, Cheng, Yang, Yang, and Xiong]{hwang2022tutel}
Changho Hwang, Wei Cui, Yifan Xiong, Ziyue Yang, Ze~Liu, Han Hu, Zilong Wang,
  Rafael Salas, Jithin Jose, Prabhat Ram, HoYuen Chau, Peng Cheng, Fan Yang,
  Mao Yang, and Yongqiang Xiong.
\newblock Tutel: Adaptive mixture-of-experts at scale.
\newblock In D.~Song, M.~Carbin, and T.~Chen (eds.), \emph{Machine Learning and
  Systems}, volume~5, pp.\  269--287, 2023{\natexlab{a}}.

\bibitem[Hwang et~al.(2023{\natexlab{b}})Hwang, Cui, Xiong, Yang, Liu, Hu,
  Wang, Salas, Jose, Ram, et~al.]{tutel}
Changho Hwang, Wei Cui, Yifan Xiong, Ziyue Yang, Ze~Liu, Han Hu, Zilong Wang,
  Rafael Salas, Jithin Jose, Prabhat Ram, et~al.
\newblock Tutel: Adaptive mixture-of-experts at scale.
\newblock \emph{Machine Learning and Systems}, 5:\penalty0 269--287,
  2023{\natexlab{b}}.

\bibitem[Jacobs et~al.(1991)Jacobs, Jordan, Nowlan, and
  Hinton]{jacobs1991adaptive}
Robert~A Jacobs, Michael~I Jordan, Steven~J Nowlan, and Geoffrey~E Hinton.
\newblock Adaptive mixtures of local experts.
\newblock \emph{Neural computation}, 3\penalty0 (1):\penalty0 79--87, 1991.

\bibitem[Kenton \& Toutanova(2019)Kenton and Toutanova]{kenton:19bert}
Jacob Devlin Ming-Wei~Chang Kenton and Lee~Kristina Toutanova.
\newblock {BERT}: Pre-training of deep bidirectional transformers for language
  understanding.
\newblock In \emph{NAACL-HLT}, volume~1, pp.\  4171--4186, 2019.

\bibitem[Komatsuzaki et~al.(2022)Komatsuzaki, Puigcerver, Lee-Thorp, Ruiz,
  Mustafa, Ainslie, Tay, Dehghani, and Houlsby]{komatsuzaki2022sparse}
Aran Komatsuzaki, Joan Puigcerver, James Lee-Thorp, Carlos~Riquelme Ruiz, Basil
  Mustafa, Joshua Ainslie, Yi~Tay, Mostafa Dehghani, and Neil Houlsby.
\newblock Sparse upcycling: Training mixture-of-experts from dense checkpoints.
\newblock \emph{arXiv:2212.05055}, 2022.

\bibitem[Lepikhin et~al.(2020)Lepikhin, Lee, Xu, Chen, Firat, Huang, Krikun,
  Shazeer, and Chen]{lepikhin2020gshard}
Dmitry Lepikhin, HyoukJoong Lee, Yuanzhong Xu, Dehao Chen, Orhan Firat, Yanping
  Huang, Maxim Krikun, Noam Shazeer, and Zhifeng Chen.
\newblock Gshard: Scaling giant models with conditional computation and
  automatic sharding.
\newblock \emph{arXiv:2006.16668}, 2020.

\bibitem[Lin et~al.(2024)Lin, Shrivastava, Luo, Iyer, Lewis, Ghosh,
  Zettlemoyer, and Aghajanyan]{lin2024moma}
Xi~Victoria Lin, Akshat Shrivastava, Liang Luo, Srinivasan Iyer, Mike Lewis,
  Gargi Ghosh, Luke Zettlemoyer, and Armen Aghajanyan.
\newblock Moma: Efficient early-fusion pre-training with mixture of
  modality-aware experts.
\newblock \emph{arXiv:2407.21770}, 2024.

\bibitem[Liu et~al.(2022)Liu, Mao, Wu, Feichtenhofer, Darrell, and
  Xie]{liu2022convnet}
Zhuang Liu, Hanzi Mao, Chao-Yuan Wu, Christoph Feichtenhofer, Trevor Darrell,
  and Saining Xie.
\newblock A convnet for the 2020s.
\newblock In \emph{IEEE/CVF Conference on Computer Vision and Pattern
  Recognition}, pp.\  11976--11986, 2022.

\bibitem[Lou et~al.(2021)Lou, Xue, Zheng, and You]{lou2021cross}
Yuxuan Lou, Fuzhao Xue, Zangwei Zheng, and Yang You.
\newblock Cross-token modeling with conditional computation.
\newblock \emph{arXiv:2109.02008}, 2021.

\bibitem[Mittal et~al.(2022)Mittal, Bengio, and Lajoie]{mittal2022modular}
Sarthak Mittal, Yoshua Bengio, and Guillaume Lajoie.
\newblock Is a modular architecture enough?
\newblock \emph{Advances in Neural Information Processing Systems},
  35:\penalty0 28747--28760, 2022.

\bibitem[Mohammed et~al.(2022)Mohammed, Liu, and Raffel]{mohammed2022models}
Muqeeth Mohammed, Haokun Liu, and Colin Raffel.
\newblock Models with conditional computation learn suboptimal solutions.
\newblock In \emph{I Can't Believe It's Not Better Workshop: Understanding Deep
  Learning Through Empirical Falsification}, 2022.

\bibitem[Mustafa et~al.(2022)Mustafa, Riquelme, Puigcerver, Jenatton, and
  Houlsby]{mustafa2022multimodal}
Basil Mustafa, Carlos Riquelme, Joan Puigcerver, Rodolphe Jenatton, and Neil
  Houlsby.
\newblock Multimodal contrastive learning with {LIMoE}: the language-image
  mixture of experts.
\newblock \emph{arXiv:2206.02770}, 2022.

\bibitem[Puigcerver et~al.(2024)Puigcerver, Ruiz, Mustafa, and
  Houlsby]{puigcerver2023sparse}
Joan Puigcerver, Carlos~Riquelme Ruiz, Basil Mustafa, and Neil Houlsby.
\newblock From sparse to soft mixtures of experts.
\newblock In \emph{12th International Conference on Learning Representations},
  2024.

\bibitem[Recht et~al.(2019)Recht, Roelofs, Schmidt, and
  Shankar]{recht2019ImageNet}
Benjamin Recht, Rebecca Roelofs, Ludwig Schmidt, and Vaishaal Shankar.
\newblock Do {ImageNet} classifiers generalize to {ImageNet} ?
\newblock In \emph{International Conference on Machine Learning}, pp.\
  5389--5400, 2019.

\bibitem[Riquelme et~al.(2021)Riquelme, Puigcerver, Mustafa, Neumann, Jenatton,
  Susano~Pinto, Keysers, and Houlsby]{riquelme2021scaling}
Carlos Riquelme, Joan Puigcerver, Basil Mustafa, Maxim Neumann, Rodolphe
  Jenatton, Andr{\'e} Susano~Pinto, Daniel Keysers, and Neil Houlsby.
\newblock Scaling vision with sparse mixture of experts.
\newblock \emph{Advances in Neural Information Processing Systems},
  34:\penalty0 8583--8595, 2021.

\bibitem[Russakovsky et~al.(2015)Russakovsky, Deng, Su, Krause, Satheesh, Ma,
  Huang, Karpathy, Khosla, Bernstein, Berg, and
  Fei-Fei]{russakovsky:imagenet:15}
Olga Russakovsky, Jia Deng, Hao Su, Jonathan Krause, Sanjeev Satheesh, Sean Ma,
  Zhiheng Huang, Andrej Karpathy, Aditya Khosla, Michael Bernstein,
  Alexander~C. Berg, and Li~Fei-Fei.
\newblock {ImageNet Large Scale Visual Recognition Challenge}.
\newblock \emph{International Journal of Computer Vision}, 115\penalty0
  (3):\penalty0 211--252, 2015.

\bibitem[Shazeer et~al.(2017)Shazeer, Mirhoseini, Maziarz, Davis, Le, Hinton,
  and Dean]{shazeer2017outrageously}
Noam Shazeer, Azalia Mirhoseini, Krzysztof Maziarz, Andy Davis, Quoc Le,
  Geoffrey Hinton, and Jeff Dean.
\newblock Outrageously large neural networks: The sparsely-gated
  mixture-of-experts layer.
\newblock \emph{arXiv:1701.06538}, 2017.

\bibitem[Strubell et~al.(2019)Strubell, Ganesh, and McCallum]{strubell:19}
Emma Strubell, Ananya Ganesh, and Andrew McCallum.
\newblock Energy and policy considerations for deep learning in nlp.
\newblock \emph{arXiv:1906.02243}, 2019.

\bibitem[Tolstikhin et~al.(2021)Tolstikhin, Houlsby, Kolesnikov, Beyer, Zhai,
  Unterthiner, Yung, Steiner, Keysers, Uszkoreit, et~al.]{tolstikhin2021mlp}
Ilya~O Tolstikhin, Neil Houlsby, Alexander Kolesnikov, Lucas Beyer, Xiaohua
  Zhai, Thomas Unterthiner, Jessica Yung, Andreas Steiner, Daniel Keysers,
  Jakob Uszkoreit, et~al.
\newblock {MLP-Mixer}: An all-{MLP} architecture for vision.
\newblock \emph{Advances in neural information processing systems},
  34:\penalty0 24261--24272, 2021.

\bibitem[Touvron et~al.(2021)Touvron, Cord, Douze, Massa, Sablayrolles, and
  J{\'e}gou]{touvron2021training}
Hugo Touvron, Matthieu Cord, Matthijs Douze, Francisco Massa, Alexandre
  Sablayrolles, and Herv{\'e} J{\'e}gou.
\newblock Training data-efficient image transformers \& distillation through
  attention.
\newblock In \emph{International conference on machine learning}, pp.\
  10347--10357. PMLR, 2021.

\bibitem[Touvron et~al.(2022)Touvron, Cord, and J{\'e}gou]{touvron2022deit}
Hugo Touvron, Matthieu Cord, and Herv{\'e} J{\'e}gou.
\newblock Deit {III}: Revenge of the {ViT}.
\newblock In \emph{17th European Conference Computer Vision}, pp.\  516--533,
  2022.

\bibitem[Tu et~al.(2022)Tu, Talebi, Zhang, Yang, Milanfar, Bovik, and
  Li]{tu2022maxvit}
Zhengzhong Tu, Hossein Talebi, Han Zhang, Feng Yang, Peyman Milanfar, Alan
  Bovik, and Yinxiao Li.
\newblock Maxvit: Multi-axis vision transformer.
\newblock In \emph{European Conference on Computer Vision}, pp.\  459--479.
  Springer, 2022.

\bibitem[Vaswani et~al.(2017)Vaswani, Shazeer, Parmar, Uszkoreit, Jones, Gomez,
  Kaiser, and Polosukhin]{vaswani2017attention}
Ashish Vaswani, Noam Shazeer, Niki Parmar, Jakob Uszkoreit, Llion Jones,
  Aidan~N Gomez, Lukasz Kaiser, and Illia Polosukhin.
\newblock Attention is all you need.
\newblock In I.~Guyon, U.~Von Luxburg, S.~Bengio, H.~Wallach, R.~Fergus,
  S.~Vishwanathan, and R.~Garnett (eds.), \emph{Advances in Neural Information
  Processing Systems}, volume~30, 2017.

\bibitem[Wang et~al.(2019)Wang, Ge, Lipton, and Xing]{wang2019learning}
Haohan Wang, Songwei Ge, Zachary Lipton, and Eric~P Xing.
\newblock Learning robust global representations by penalizing local predictive
  power.
\newblock In \emph{Advances in Neural Information Processing Systems}, pp.\
  10506--10518, 2019.

\bibitem[Yu et~al.(2022)Yu, Wang, Vasudevan, Yeung, Seyedhosseini, and
  Wu]{yu2022coca}
Jiahui Yu, Zirui Wang, Vijay Vasudevan, Legg Yeung, Mojtaba Seyedhosseini, and
  Yonghui Wu.
\newblock Coca: Contrastive captioners are image-text foundation models.
\newblock \emph{arXiv:2205.01917}, 2022.

\bibitem[Yun et~al.(2019)Yun, Han, Oh, Chun, Choe, and Yoo]{yun:19:ccv}
Sangdoo Yun, Dongyoon Han, Seong~Joon Oh, Sanghyuk Chun, Junsuk Choe, and
  Youngjoon Yoo.
\newblock Cutmix: Regularization strategy to train strong classifiers with
  localizable features.
\newblock In \emph{IEEE/CVF International Conference on Computer Vision}, 2019.

\bibitem[Zhai et~al.(2022)Zhai, Kolesnikov, Houlsby, and
  Beyer]{zhai2022scaling}
Xiaohua Zhai, Alexander Kolesnikov, Neil Houlsby, and Lucas Beyer.
\newblock Scaling vision transformers.
\newblock In \emph{IEEE/CVF Conference on Computer Vision and Pattern
  Recognition}, pp.\  12104--12113, 2022.

\bibitem[Zhang et~al.(2018)Zhang, Cisse, Dauphin, and
  Lopez-Paz]{zhang2017mixup}
Hongyi Zhang, Moustapha Cisse, Yann~N. Dauphin, and David Lopez-Paz.
\newblock mixup: Beyond empirical risk minimization.
\newblock In \emph{International Conference on Learning Representations}, 2018.

\bibitem[Zhong et~al.(2020)Zhong, Zheng, Kang, Li, and Yang]{zhong2020random}
Zhun Zhong, Liang Zheng, Guoliang Kang, Shaozi Li, and Yi~Yang.
\newblock Random erasing data augmentation.
\newblock \emph{AAAI Conference on Artificial Intelligence}, 34\penalty0
  (07):\penalty0 13001--13008, 2020.

\bibitem[Zoph et~al.(2022)Zoph, Bello, Kumar, Du, Huang, Dean, Shazeer, and
  Fedus]{zoph2022designing}
Barret Zoph, Irwan Bello, Sameer Kumar, Nan Du, Yanping Huang, Jeff Dean, Noam
  Shazeer, and William Fedus.
\newblock Designing effective sparse expert models.
\newblock \emph{arXiv:2202.08906}, 2022.

\end{thebibliography}

\clearpage
\appendix
\clearpage

\FloatBarrier
\section{Detailed Hyperparameters}\label{sec:detailedhypers}
\change{We train all models with distributed data parallel~(DDP) across devices and expert-parallelism within layers using Tutel~\citep{tutel} for efficient \moe{} implementation.}
\FloatBarrier
\begin{table}[!ht]
    \centering    
    {    
    \begin{tabular}{l|cc|cc}
    \toprule
    \multicolumn{1}{c}{Hyperparameter} & \multicolumn{2}{c}{ViT \cite{touvron2022deit}} & \multicolumn{2}{c}{\ConvNeXt{} \cite{liu2022convnet}} \\
    \multicolumn{1}{c}{Dataset} & IN1k & IN22k  & IN1k & IN21k \\
    \midrule
    Batch size & 2048 & 2048 & 2048 & 2048\\
    Optimizer & LAMB & LAMB & AdamW & AdamW\\
    LR & $3\cdot10^{-3}$ & $2.5\cdot10^{-3}$ & $2.5\cdot10^{-3}$ & $2.5\cdot10^{-3}$\\
    LR decay & cosine & cosine & cosine & cosine\\ 
    Weight decay & 0.03 & 0.02 & 0.05 & 0.05\\
    Expert Weight d. & 0.06 & 0.04 & 0.05 & 0.05\\
    Warmup epochs & 5 & 5 & 20 & 5\\
    Load balancing & 0.01 & 0.01 & 0.01 & 0.01 \\
    \cmidrule{1-5}
    Label smoothing $\epsilon$ & 0.0 & 0.1 & 0.1 & 0.1\\
    Stoch.Depth & \cmark & \cmarkg & \cmarkg & \cmarkg\\
    RepeatedAug & \cmark & \xmarkg & \xmarkg & \xmarkg\\
    \cmidrule{1-5}
    H.flip & \cmark & \cmark & \cmark & \cmark\\
    RRC & \xmarkg & \cmark & \xmarkg & \xmarkg\\
    Rand Augment & \xmarkg & \xmarkg & \cmark & \cmark\\ 
    3Augment & \cmark & \cmark & \xmarkg & \xmarkg\\
    Color Jitter & 0.3 & 0.3 & \xmarkg & \xmarkg\\
    Mixup alpha & 0.8 & 0.0 & 0.8 & 0.0 \\
    Cutmix alpha & 1.0 & 1.0 & 1.0 & 1.0\\
    Erasing prob. & \xmarkg & \xmarkg & 0.25 & \xmarkg\\
    \cmidrule{1-5}
    Test crop ratio & 1.0 & 1.0 & 0.875 & 0.875\\
    \cmidrule{1-5}
    Loss & BCE & CE & CE & CE\\
    \bottomrule
    \end{tabular}   
    }
\end{table}

\begin{table}[!ht]
  \centering    
    {
    \caption{Summary of our training procedures with \imagenetonek{} (IN1k) and \imagenettwok{} (IN21k).}\label{tab:hparams}
    \begin{tabular}{l|c|c}
    \toprule
    \multicolumn{1}{c}{Architecture} & \imagenetonek{} & \imagenettwok{}{}\\
    \midrule
      ViT-(S/B) & 0.1 / 0.2 & 0.0 / 0.1\\
      ConvNext-(T/S/B) &  0.1 / 0.3 / 0.6 & 0.0 / 0.0 / 0.1\\
      ConvNext-(T/B) \textit{iso.}&  0.1 / 0.5 & -\\
    \bottomrule
    \end{tabular}
    }    
\end{table}
\FloatBarrier


\clearpage


\clearpage
\FloatBarrier

\section{\moe{} vs Dense Table}\label{app:moedense}
\FloatBarrier
\begin{table}[!htpb]    
    \centering\normalsize{%
    \caption{Experiments with pretraining on \imagenettwok{}, JFT-300M or JFT-3B: performance with/without \moe{}. In the case of ConvNeXt-T, we get 0.6 improvement, while increasing the number of parameters, but almost not the number of per-sample parameters. For ConvNeXt-S, there is a 0.3 improvement. There is an improvement for SwinV2 models, but these were starting lower. Overall, for strong/big models, there is no clear improvement. Results without citations correspond to our work. V-MoE results marked with a star ($^*$) are results taken from their GitHub (\hurl{tinyurl.com/yb8fze5u}). See~\cref{fig:max_ipvt} for an overview. For every model size examined, we present the accuracy metrics obtained after similar numbers of training iterations. Accuracy (Acc.) refers to Top-1 accuracy.}\label{tab:21k:moe}%
    \begin{tabular}{lcccc}
        \toprule
         \multicolumn{1}{c}{\multirow{2}{*}{Architecture}} & \multicolumn{1}{c}{\multirow{2}{*}{Pre-train dataset}} &\verb|#|Params & {\small Per samples} & IN-1K\\
          & & ($\times10^6$) & \verb|#|Params$_{act}$  & Acc.\\ 
         \midrule
        ViT-s/32\citep{riquelme2021scaling} & JFT-300M & 36.5 & 36.5 & 73.73\\
        ViT-s/32-32, Last 2\citep{riquelme2021scaling} & JFT-300M & 166.7 & $\approx 55$ & 77.10\\
        ViT-s/32-32, Every 2\citep{riquelme2021scaling} & JFT-300M & 296.9 & $\approx 70$ & 77.10\\
        \hline
        \rowcolor{gray!15}ViT-S/16 \citep{touvron2022deit} & \imagenettwok{} & 22.0 & 22.0 & 82.6\\ 
        \rowcolor{gray!15} ViT-S/16-8 Every 2 Top 2 & \imagenettwok{} & 71.7.6 & 33.1 & 83.0\\ 
        \hline
        ConvNeXt-T \citep{liu2022convnet} & \imagenettwok{} & 28.6 & 28.6 & 82.9\\
        ConvNeXt-T-8 Last 2 Top 1 & \imagenettwok{} & 70.0 & 28.7 & \textbf{83.5}\\
        \hline
        \rowcolor{gray!15}  SwinV2-S \citep{hwang2022tutel} & \imagenettwok{} & 65.8 & - & 83.5\\
        \rowcolor{gray!15}  SwinV2-S-8 \citep{hwang2022tutel} & \imagenettwok{} & 173.3 & 65.8 & 84.5\\
        \rowcolor{gray!15}  SwinV2-S-16 \citep{hwang2022tutel} & \imagenettwok{} & 296.1 & 65.8 & \textbf{84.9}\\
        \rowcolor{gray!15} SwinV2-S-32 \citep{hwang2022tutel} & \imagenettwok{} & 296.1 & 65.8 & 84.7\\
        \hline
        ConvNeXt-S  \citep{liu2022convnet} & \imagenettwok{} & 50.3 & - & 84.6 \\
        ConvNeXt-S-8 Last 2 Top 1 & \imagenettwok{} & 91.6 & 50.3 & \textbf{84.9}\\
        \hline
        \rowcolor{gray!15} ViT-B/16 \citep{touvron2022deit} & \imagenettwok{} & 86.6 & 86.6 & 85.2\\
        \rowcolor{gray!15} ViT-B/16, Every 2, Top 2& \imagenettwok{} & 284.9 & 129.9 & 85.2\\
        \rowcolor{gray!15} ViT-B/16$\uparrow 384$ \citep{touvron2022deit} & \imagenettwok{} & 86.6 & 86.6 & 86.7\\
        \rowcolor{gray!15} ViT-B/16 \citep{zhai2022scaling} & JFT-300M & 86.6 & 86.6 & 84.9\\
        \rowcolor{gray!15} ViT-B/16 \citep{riquelme2021scaling} & JFT-300M & 100.5 & 100.5 & 84.15\\
        \hline
        V-MoE-B/16-32, Every 2 \citep{riquelme2021scaling} & JFT-300M & 979.0 & $\approx 200$ & 85.3\\
        V-MoE-B/16-32, Last 2 \citep{riquelme2021scaling} & JFT-300M & 393.3 & $\approx 110$ & \textbf{85.4}\\
        \hline
        \rowcolor{gray!15} SwinV2-B \citep{hwang2022tutel} & \imagenettwok{} & 109.3 & 109.3 & 85.1\\
        \rowcolor{gray!15} SwinV2-B-8 \citep{hwang2022tutel} & \imagenettwok{} & 300.3 & 109.3 & 85.3\\
        \rowcolor{gray!15} SwinV2-B-16 \citep{hwang2022tutel} & \imagenettwok{} & 518.7 & 109.3 & \textbf{85.5}\\
        \rowcolor{gray!15} SwinV2-B-32 \citep{hwang2022tutel} & \imagenettwok{} & 955.3 & 109.3 & \textbf{85.5}\\
        \hline
        ConvNeXt-B \citep{liu2022convnet} & \imagenettwok{} & 88.6 & 88.6 & \textbf{85.8}\\
        ConvNeXt-B-8 & \imagenettwok{} & 162.0 & 88.6 & 85.7\\
        \hline
        \rowcolor{gray!15} ViT-L/16 \citep{touvron2022deit} & \imagenettwok{} & 304.4 & 304.4 & 87.0\\
        \rowcolor{gray!15} ViT-L/16 \citep{zhai2022scaling} & JFT-300M & 323.1 & 323.1 & \textbf{87.7}\\
        \rowcolor{gray!15} ViT-L/16 \citep{riquelme2021scaling} & JFT-300M & 323.1 & 323.1 & 87.1\\
        \hline
        V-MoE-L/16-32, Every 2 \citep{riquelme2021scaling} & JFT-300M & 3446.0 & $\approx 600$ & 87.4\\
        V-MoE-L/16-32, Last 2 \citep{riquelme2021scaling} & JFT-300M & 843.6 & $\approx 380$ & 87.5\\
        \hline
        \rowcolor{gray!15} SoViT/14\citep{alabdulmohsin2023getting} & JFT-3B & 400 & 400 & 90.3\\
        \rowcolor{gray!15} ViT-g/14\citep{zhai2022scaling} & JFT-3B & 1B & 1B & \textbf{90.45}\\
        \rowcolor{gray!15} V-MoE/14 \citep{riquelme2021scaling} & JFT-3B & 15B & $\approx 1$B & 90.35\\
        \bottomrule
    \end{tabular}
}
\end{table}

\clearpage
\section{Ade20k}
\begin{table*}[!ht]
    \centering
    \caption{ImageNet-1K trained models}\label{tab:1k:reduced}
    \begin{tabular}{lcccc}
        \hline
        \multicolumn{1}{c}{\multirow{2}{*}{Backbone}} & \multirow{2}{*}{Input crop.} & \verb|#|Params & {\small per sample} & Single scale \\
        & ($\times10^6$) & \verb|#|Params  &  & mIoU\\ 
        \hline
        \multicolumn{5}{c}{\bf ImageNet-1K trained models}\\
        \hline
        ConvNeXt-T~\cite{liu2022convnet} & $512^2$ &28.6 & - &  46.0 \\
        \rowcolor{gray!15} ConvNeXt-T-4 Last 2 Top 1 & $512^2$ & 34.5 & 25.6 & 46.0 \\
        \hline
        ConvNeXt-S \cite{liu2022convnet} & $512^2$ & 50 & - &  48.7\\
        \rowcolor{gray!15} ConvNeXt-S-4 Last 2 Top 1 & $512^2$ & 56.1 & 47.3 & 48.38\\
        \hline
        ConvNeXt-B \cite{liu2022convnet}  & $512^2$ & 88.6 & - & 49.1\\
        \rowcolor{gray!15} ConvNeXt-B-4 Last 2 Top 1 & $512^2$ & 99.1 & 83.4 & 48.8 \\
        \hline
        ViT-S \cite{touvron2022deit}  & $512^2$ & 88.6 & - & 45.8\\
        \rowcolor{gray!15} ViT-S-8 Every 2 Top 2 & $512^2$ & 99.1 & 83.4 & 45.4\\
        \hline
        ViT-B \cite{touvron2022deit}  & $512^2$ & 88.6 & - & 49.0 \\
        \rowcolor{gray!15} ViT-B-8 Every 2 Top 2 & $512^2$ & 99.1 & 83.4 & 48.3\\
        \hline
        \multicolumn{5}{c}{\bf ImageNet-22K trained models}\\
        \hline
        ConvNeXt-T~\cite{liu2022convnet} & $640^2$ &28.6 & - &  48.6 \\
        \rowcolor{gray!15} ConvNeXt-T-4 Last 2 Top 1 & $640^2$ & 34.5 & 25.6 & 48.9 \\
        \hline
        ConvNeXt-S \cite{liu2022convnet} & $640^2$ & 50 & - &  51.0\\
        \rowcolor{gray!15} ConvNeXt-S-4 Last 2 Top 1 & $640^2$ & 56.1 & 47.3 & 50.8\\
        \hline
        ConvNeXt-B \cite{liu2022convnet}  & $640^2$ & 88.6 & - & 52.6\\
        \rowcolor{gray!15} ConvNeXt-B-4 Last 2 Top 1 & $640^2$ & 99.1 & 83.4 & 52.0 \\
        \hline
        ViT-S \cite{touvron2022deit}  & $640^2$ & 88.6 & - & 48.4\\
        \rowcolor{gray!15} ViT-S-8 Every 2 Top 2 & $640^2$ & 99.1 & 83.4 & 48.7\\
        \hline
        ViT-B \cite{touvron2022deit}  & $640^2$ & 88.6 & - &  52.8\\
        \rowcolor{gray!15} ViT-B-8 Every 2 Top 2 & $640^2$ & 99.1 & 83.4 & 52.4 \\
        \hline
    \end{tabular}    
\end{table*}
%

\end{document}